\documentclass[journal]{IEEEtran}
\usepackage{times}
\usepackage{epsfig}
\usepackage{graphicx}
\usepackage{amsmath}
\usepackage{amssymb}
\usepackage{array}
\usepackage{makecell}
\usepackage{multirow}
\usepackage{animate}
\usepackage{wrapfig}
\usepackage{enumitem}
\usepackage{float}
\newcolumntype{P}[1]{>{\centering\arraybackslash}p{#1}}
\usepackage{caption}
\usepackage[pagebackref=true,breaklinks=true,letterpaper=true,colorlinks,bookmarks=false]{hyperref}

\ifCLASSINFOpdf
\else
\fi
\begin{document}
%
\title{Deep Spatial Transformation for Pose-Guided \\ Person Image Generation and Animation}
%
%
%

\author{Yurui~Ren,
        Ge~Li,
        Shan Liu,
        ~and~Thomas H. Li
\thanks{Y. Ren, and G. Li are with School of Electronic and Computer Engineering, Peking University, Shenzhen 518055, China (e-mail: yrren@pku.edu.cn; geli@ece.pku.edu.cn).}
\thanks{S. Liu is with Tencent America, Palo Alto, CA 94301 USA (e-mail: shanl@tencent.com).}
\thanks{T. H. Li is with Advanced Institute of Information Technology, Peking University, HangZhou 311215, China (e-mail: tli@aiit.org.cn).}
}

\maketitle

\begin{abstract}
Pose-guided person image generation and animation aim to transform a source person image to target poses. These tasks require spatial manipulation of source data. 
However, Convolutional Neural Networks are limited by the lack of ability to spatially transform the inputs. In this paper, we propose a differentiable global-flow local-attention framework to reassemble the inputs at the feature level. This framework first estimates global flow fields between sources and targets. Then, corresponding local source feature patches are sampled with content-aware local attention coefficients.
We show that our framework can spatially transform the inputs in an efficient manner. Meanwhile, we further model the temporal consistency for the person image animation task to generate coherent videos. 
The experiment results of both image generation and animation tasks demonstrate the superiority of our model. Besides, additional results of novel view synthesis and face image animation show that our model is applicable to other tasks requiring spatial transformation. The source code of our project is available at \url{https://github.com/RenYurui/Global-Flow-Local-Attention}.



\end{abstract}


%
\IEEEpeerreviewmaketitle

\section{Introduction}
\begin{figure}[t]
\begin{center}
\includegraphics[width=1\linewidth]{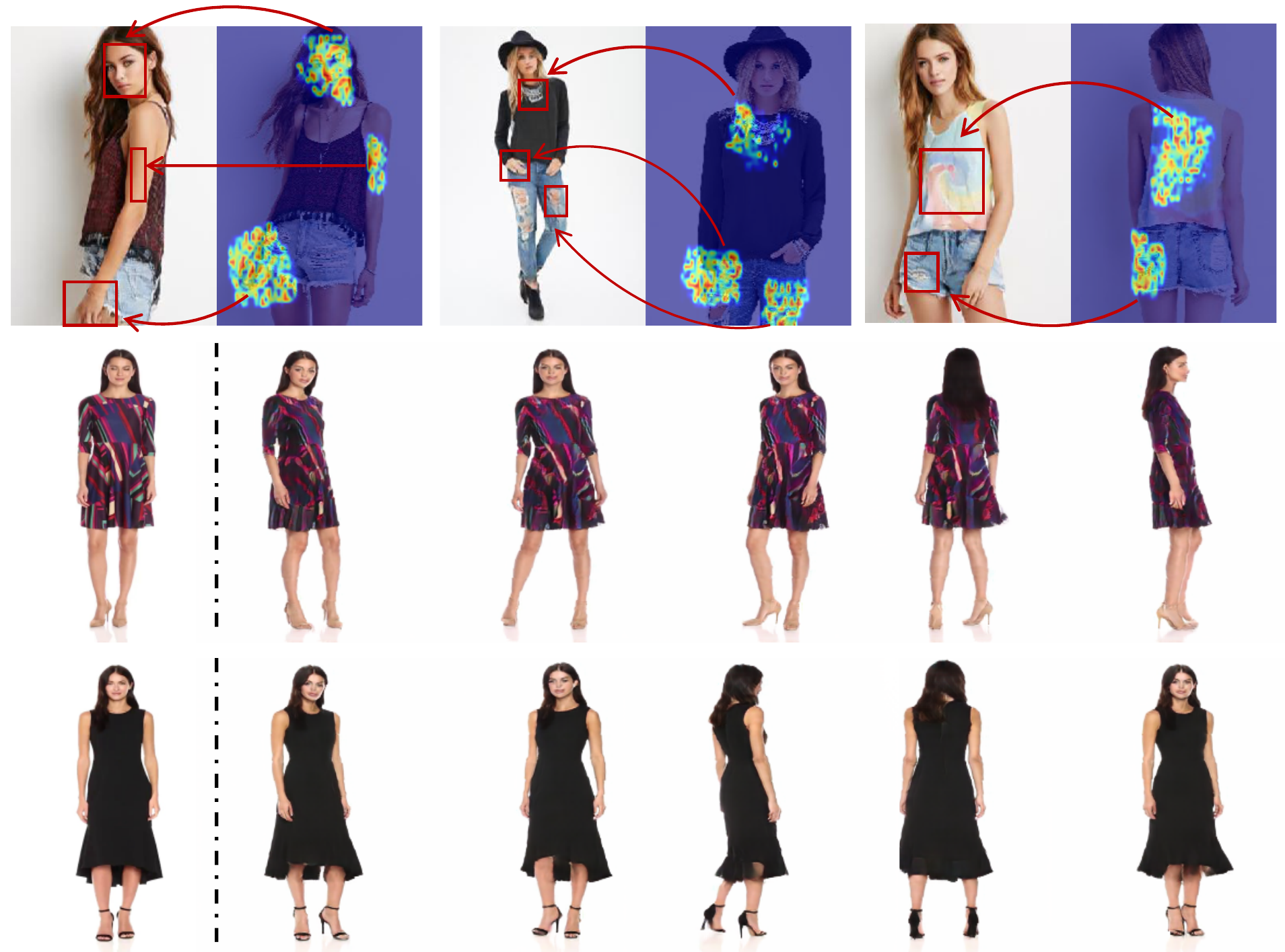}
\end{center}
   \caption{Illustration of pose-guided person image generation and animation. We show the person image generation task in the first row. For each image pair, the left image is the generated result of our model, while the right image is the input source image. The arrows indicate the data spatial transformation. The second and third rows contain results of the person image animation task. The leftmost image of each row is the source image and the others are the generated results of our model.}
\label{fig:introduction}
\end{figure}

%
%
%
%
\IEEEPARstart{I}{n} 
this paper, we deal with the conditional generation task where the target images are the spatial deformation versions of the source images. Such deformation can be caused by object motions or viewpoint changes. This task is the core of many image/video generation problems. For example, pose-guided person image generation~\cite{ma2017pose,siarohin2018deformable,song2019unsupervised,zhu2019progressive} transforms a person image from a source pose to a target pose while retaining the source appearance details. The corresponding pose-guided image animation task~\cite{wang2019few,siarohin_firstorder,Siarohin_2019_CVPR,liu2019liquid} further models the temporal consistency and generates a video from a still source image according to a driving target pose sequence. As illustrated in Figure~\ref{fig:introduction},
these tasks can be tackled by reasonably reassembling the input data in the spatial domain.

However, Convolutional Neural Networks (CNNs) lack the ability to spatially transform the input features in a parameter efficient manner. One important property of CNNs is the equivariance to transformation~\cite{goodfellow2016deep}, which means that if the input spatially shifts, then the output shifts in the same way. This property can benefit tasks requiring reasoning about images such as segmentation~\cite{girshick2014rich,he2017mask}, detection~\cite{simonyan2014two}, \emph{etc}. However, it limits the networks by the lack of ability to deal with the deformable-object generation task which requires spatially rearranging the input data.  
In order to enable spatial transformation capabilities of CNNs, Spatial Transformer Networks (STN)~\cite{jaderberg2015spatial} introduces a Spatial Transformer module to standard neural networks. This module regresses transformation parameters and warps the input features using a global affine transformation. However, the global affine transformation is not sufficient in representing the complex deformations of non-rigid objects.

The attention mechanism~\cite{vaswani2017attention,zhang2018self} is able to transform information beyond local regions. It gives networks the ability to build long-term dependencies by allowing networks to use non-local features. It has emerged as an effective technique for many tasks such as natural language processing~\cite{vaswani2017attention}, image recognition~\cite{wang2018non,hu2019local}, and image generation~\cite{zhang2018self}. However, for spatial transformation tasks in which target images are the deformation results of source images, each output position has a clear one-to-one relationship with the source position. Therefore, each output feature is only related to a local region of the source features. In other words, the attention coefficient matrix between the source and target should be a sparse matrix instead of a dense matrix.

Flow-based operation forces the attention coefficient matrix to be a sparse matrix by sampling a very local source patch for each output position. It predicts 2D coordinate offsets for the target features specifying the sampling source positions. 
However, networks struggle to find reasonable sampling locations when warping the inputs at the feature level~\cite{yu2018generative,ren2020deep}.
Possible explanations for this phenomenon are that: $(1)$ The input features and flow fields change simultaneously during the training stage. Their parameter update processes are mutually constrained, which means that the input features cannot obtain reasonable gradients without correct flow fields and vice versa. $(2)$ The commonly used Bilinear sampling method provides poor gradient propagation~\cite{jiang2019linearized,ren2019structureflow}; In order to obtain meaningful flow fields, some flow-based methods~\cite{zhou2016view,fischer2015flownet} warp input data at the pixel level. However, this operation limits the networks to be unable to generate new content. Meanwhile, large motions are difficult to be extracted due to the requirement of generating full-resolution flow fields~\cite{ranjan2017optical}. Some methods warp the input at the feature level by pre-calculating the flow fields using additional 3D models~\cite{liu2019liquid} or generate dense flow fields from sparse key point representation~\cite{Siarohin_2019_CVPR}. However, they do not solve the problems in a straightforward manner, which leads to an insufficient transformation representation capability.

\begin{figure*}[t]
\begin{center}
\includegraphics[width=0.97\linewidth]{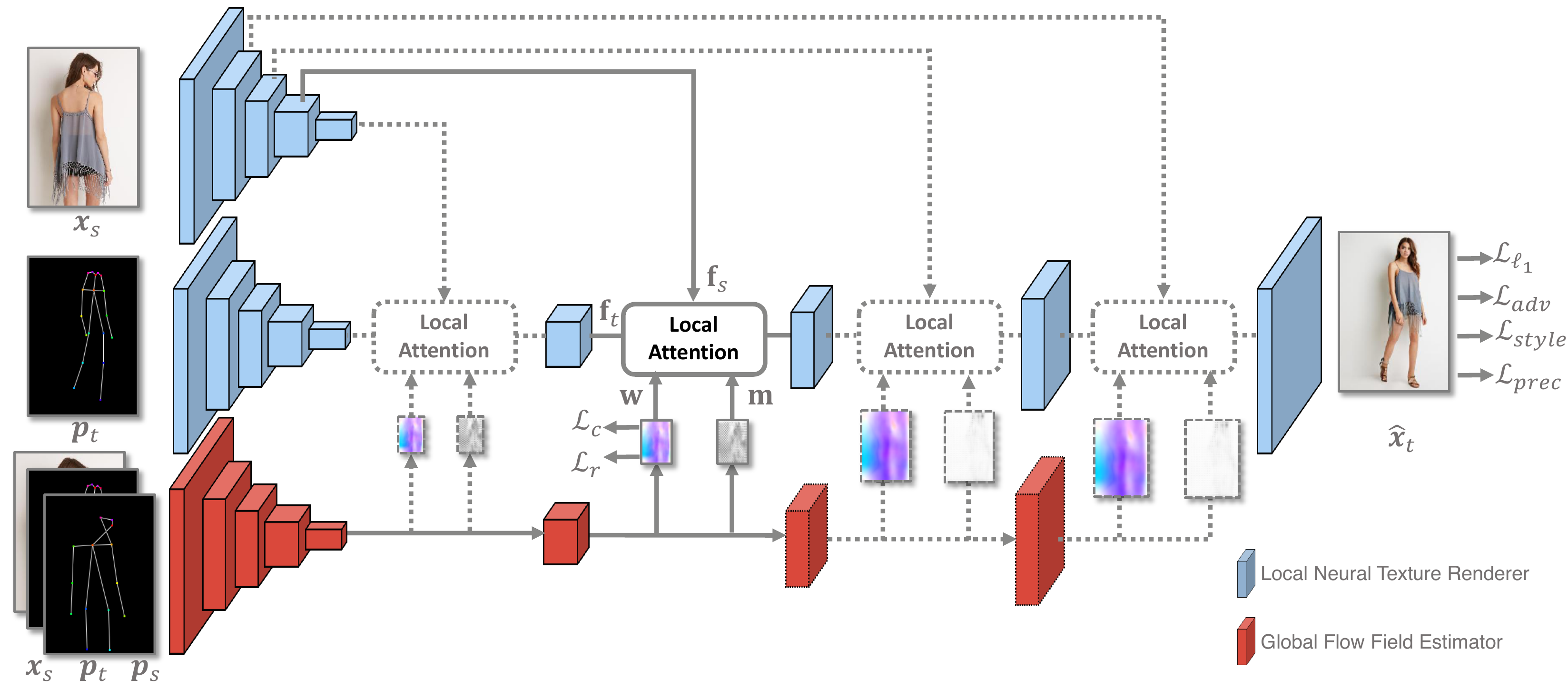}
\end{center}
   \caption{Overview of our GFLA model. The Global Flow Field Estimator is used to generate flow fields. The Local Neural Texture Renderer yields results by spatially transforming the source features using local attention. Dotted lines indicate that our local attention module can be used at different scales.}
 
\label{fig:OurModel}
\end{figure*}

In this paper, we propose a differentiable \textit{Global-Flow Local-Attention (GFLA) framework} to solve the problems. Our framework can enable CNNs to reasonably sample and reassemble source features without using any labeled flow fields.
The architecture of our GFLA framework can be found in Figure~\ref{fig:OurModel}.
Specifically, our network can be divided into two parts: Global Flow Field Estimator and Local Neural Texture Renderer. The Global Flow Field Estimator is responsible for extracting the long-term dependencies between sources and targets. It estimates flow fields that assign a local source feature patch for each target position. The Local Neural Texture Renderer uses the extracted flow fields to sample the vivid source neural textures.
In order to warp sources at the feature level, we propose several targeted solutions to deal with the analyzed problems. First, a Sampling Correctness loss is proposed to constrain flow fields to sample semantically similar regions. This loss helps with the convergence by providing flow fields with additional gradients that are not related to the input source features. Then, a content-aware sampling method is proposed to avoid the poor gradient propagation of the Bilinear sampling. Experiments show that our framework is able to spatially transform the information in an efficient manner. Ablation studies demonstrate that the proposed improvements are helpful for the convergence.

The image-based pose transformation can be further extended for the pose animation task by coherently rendering an input skeleton video. However, most existing models~\cite{Siarohin_2019_CVPR,liu2019liquid,siarohin_firstorder,zablotskaia2019dwnet} directly apply image transformation methods for this task and generate each video frame independently. This operation does not take the correlations of adjacent frames into consideration, which leads to temporally inconsistent results. In order to obtain coherent results, we make additional efforts to model the temporal dynamics. 
We notice that the input skeleton sequences extracted by popular pose estimation models~\cite{cao2018openpose,li2018crowdpose} are always inconsistent.
Since these models predict result poses in an image-based manner and do not consider the temporal information of videos, obvious noise can be observed in their results.
Therefore, we propose a Motion Extraction Network to extract clean skeleton sequences from the corresponding noise data. Meanwhile, we improve our GFLA model to generate video clips in a recurrent manner. It allows our model to explicitly extract the correlations between adjacent frames. Ablation studies show that these methods can efficiently improve the final results.

We compare our model with several state-of-the-art methods over both pose-guided image generation and animation tasks. The subjective and objective experiments demonstrate the superiority of our model. Besides, we show that our model is not limited to generating person images. Additional experiments are conducted over other tasks requiring spatial transformation manipulation including novel view synthesis and face image animation. The results show the versatility of our module. The main contributions of our paper can be summarized as: 

\begin{itemize}[leftmargin=*]
  \item A GFLA model is proposed for deep spatial transformation. Experiments on the pose-guided person image generation task show that our model is able to spatially transform the source neural textures in an efficient manner.



  \item The temporal consistency is further modeled for the person image animation task. Experiments demonstrate that our simple yet efficient improvements can help the model in generating coherent results.


  \item We show the versatility of our model. Additional experiments on novel view synthesis and face image animation demonstrate that our model can be flexibly applied to other tasks requiring spatial transformation.

\end{itemize}

A preliminary version of our work has been presented in~\cite{ren2020deep}. In this journal article, we improve our work from the following aspects: (1) 
We present more in-depth analyses of our GFLA model including a more extensive ablation study to evaluate the efficacy of the components and a more thorough analysis to explain the model performance.
(2) We extent our model to tackle the person image animation task. 
A Motion Extraction Network is proposed to extract clean skeletons from noise inputs. Meanwhile, a sequential GFLA model is presented to model the correlations of the adjacent frames.
(3) We provide comprehensive ablation studies and comparison experiments to evaluate the efficacy of our animation model.

\section{Related Work}
\label{sec:related_work}

\noindent
\textbf{Pose-guided Person Image Generation.}
An overview of current monocular state-of-the-art pose-guided person image generation approaches is given in Table~\ref{tab:related_work}.
We discuss these methods in detail here.
An early attempt~\cite{ma2017pose} performs this task with a two-stage network.
It first generates a coarse image with the target pose and then refines the result in an adversarial way. Esser \emph{et al.}~\cite{esser2018variational} propose to disentangle the appearance and pose of person images. However, they ignore the spatial distribution of the original appearance, which limits the network to generate complex textures.
Siarohin \emph{et al.}~\cite{siarohin2018deformable} propose that efficient deformation operations are essential for reconstructing realistic results.  
They assume that the complex deformation between sources and targets can be well approximated by a set of local affine transformations (\emph{e.g.} arms and legs \emph{etc.}). A deformable skip connection module is introduced in their paper to spatially transform image textures.
However, the pre-defined transformation components limit the application of this method. Zhu \emph{et al.}~\cite{zhu2019progressive} propose a more flexible method by using a progressive attention module. However, information may be lost during multiple transfers, which may result in blurry details. Han \emph{et al.}~\cite{han2019clothflow} propose a method using flow-based operation for information transform. However, to ease the optimization, they warp the inputs at the pixel level, which means that further refinement networks are required to fill the holes of occlusion contents. 
Li \emph{et al.}~\cite{li2019dense} warp the inputs at the feature level, which enables the network to generate occluded contents. But their method requires additional 3D human models to calculate the ground-truth flow field labels. Our model does not require any supplementary information and obtains accurate flow fields in a self-supervised manner.

\begin{table}[t]
\centering
\setlength\extrarowheight{1pt}
\resizebox{.5\textwidth}{!}{%
\centering
\begin{tabular}{cc||c|c|c|c}
\hline
&\multirow{2}{*}{} & \multirow{2}{*}{\begin{tabular}[c]{@{}c@{}}Deformation\\ Type\end{tabular}} & \multirow{2}{*}{\begin{tabular}[c]{@{}c@{}}Flow Field\\ Label\end{tabular}} & \multirow{2}{*}{\begin{tabular}[c]{@{}c@{}}For Specific\\ Subject\end{tabular}} & \multirow{2}{*}{\begin{tabular}[c]{@{}c@{}}Temporal\\ Coherence\end{tabular}} \\
                  &                                                                             &                                                                             &                                                                                 &                                                                               \\ \hline
&&&&\\ [-8pt] \hline
\multicolumn{1}{l|}{\multirow{6}{*}{\rotatebox{90}{Generation}}}&Esser \emph{et al.}~\cite{esser2018variational}           & -                                                                           & -                                                                           & N                                                                               & N                                                                             \\ 
\multicolumn{1}{l|}{}&Siarohin \emph{et al.}~\cite{siarohin2018deformable}           & Multi-Affine Trans                                                          & -                                                                           & N                                                                               & N                                                                             \\ 
\multicolumn{1}{l|}{}&Zhu \emph{et al.}~\cite{zhu2019progressive}         & Progressive Attn                                                            & -                                                                           & N                                                                               & N                                                                             \\ 
\multicolumn{1}{l|}{}&Han \emph{et al.}~\cite{han2019clothflow}    & Pixel Flow                                                            & N                                                                           & N                                                                               & N                                                                             \\ 
\multicolumn{1}{l|}{}&Li \emph{et al.}~\cite{li2019dense}         & Feature Flow                                                                & Y                                                                           & N                                                                               & N                                                                             \\ 
\multicolumn{1}{l|}{}&Ours              & Feature Flow                                                                & N                                                                           & N                                                                               & N                                                                             \\ \hline
\multicolumn{1}{l|}{}&&&&&\\ [-8pt] \hline
\multicolumn{1}{l|}{\multirow{6}{*}{\rotatebox{90}{Animation}}}&Wang \emph{et al.}~\cite{wang2018video}               & Pixel Flow                                                                  & Y                                                                           & Y                                                                               & Y                                                                             \\ 
\multicolumn{1}{l|}{}&Chan \emph{et al.}~\cite{chan2018everybody}        & -                                                                & -                                                                           & Y                                                                               & Y                                                                             \\ 

\multicolumn{1}{l|}{}&Liu \emph{et al.}~\cite{liu2019liquid}        & Feature Flow                                                                & Y                                                                           & N                                                                               & N                                                                             \\ 
\multicolumn{1}{l|}{}&Siarohin \emph{et al.}~\cite{siarohin_firstorder}        & Feature Flow                                                                & N                                                                           & N                                                                               & N                                                                             \\ 
\multicolumn{1}{l|}{}&Wang \emph{et al.}~\cite{wang2019few}       & Pixel Flow                                                                  & Y                                                                           & N                                                                               & Y                                                                             \\ 
\multicolumn{1}{l|}{}&Ours-Animation    & Feature Flow                                                                & N                                                                           & N                                                                               & Y                                                                             \\ \hline
\end{tabular}%

}
\caption{Comparison of the state-of-the-art pose-guided person image generation and animation methods. Methods are compared from four aspects. What deformation module does the method use; Whether the flow field labels are required for training or inference; Is the model trained for a specific subject; Whether temporal coherence is explicitly enforced during training. More details can be found in Section~\ref{sec:related_work}}
\label{tab:related_work}
\end{table}

\noindent
\textbf{Pose-guided Person Image Animation.}
Taking advantage of the generation capabilities of CNNs, many papers~\cite{wang2018video,chan2018everybody,aberman2019deep} deal with this task based on conditional generative adversarial networks (CGANs). Their key idea is to train a mapping function to generate realistic images by mimicking the distribution of training sets.
However, as summarized in Table~\ref{tab:related_work}, these methods are trained to generate specific subjects \emph{i.e.}  new models are required to be trained when animating new content.
To deal with this problem, Liu \emph{et al.}~\cite{liu2019liquid} propose to extract flow fields from predicted 3D body meshes and use a liquid warping module to transform source features. However, the performance of this model is limited by the accuracy of the 3D human mesh prediction. Paper~\cite{zablotskaia2019dwnet} calculates the dense warp grids according to the UV coordinates extracted by Densepose~\cite{alp2018densepose}. Again, it relies on accurate UV coordinates to generate realistic images. Some methods predict the dense flow fields from sparse key points movements. Paper~\cite{Siarohin_2019_CVPR} and paper~\cite{siarohin_firstorder} use zeroth-order and first-order Taylor expansions to approximate the complex transformations using a set of sparse trajectories respectively.
However, these methods do not explicitly model the video temporal coherence, which may lead to inconsistent movements. 
Wang \emph{et al.}~\cite{wang2019few} propose a sequential generator to model the correlations between adjacent frames and generate coherent videos. 
Combining the advantages of the previous works, our animation model can efficiently transform the source neural textures and generate coherent results.

\noindent
\textbf{Sparse Attention in Image Generation.}
The attention mechanism~\cite{vaswani2017attention} enables networks to model long-term spatial dependencies. 
It has emerged as a powerful technique to improve the performance of the image generation tasks~\cite{zhang2018self,brock2018large}.
However, the standard dense attention module is computationally inefficient. Meanwhile, the dense connection affects networks to benefit from the image locality~\cite{daras2020your}. To mitigate these limitations, paper~\cite{child2019generating} introduces Sparse Transformers which separate the full attention operation across several steps of attention. For each step, only a subset of input positions is attended for calculation. Sparse Transformers attain better performance than dense attention with significantly fewer operations. Daras \emph{et al.}~\cite{daras2020your} propose that local sparse attention kernels introduced in Sparse Transformers are mainly designed for one-dimensional data. They introduce a new local sparse attention layer that preserves two-dimensional image locality and achieves better performance. Instead of separating the dense attention, some methods~\cite{zhang2020cross,jiang2020psgan} achieve sparse attention by controlling the sharpness of the softmax function. Our GFLA model can be seen as a type of sparse attention module, where only the flowed local patches are used for the attention coefficient calculation.


\section{Global-Flow Local-Attention \\ for Person Image Generation}
\label{image_generation}
For the pose-guided person image generation task, target images are the deformation versions of source images. Therefore, target images can be generated by spatially transforming the source images. In this section, we describe a GFLA model to efficiently warp and reassemble source neural textures. 
The architecture of our model can be found in Figure~\ref{fig:OurModel}. It can be divided into two modules: \textit{Global Flow Field Estimator} $F$ and \textit{Local Neural Texture Renderer} $G$. The Global Flow Field Estimator is responsible for estimating the global motions between sources and targets. Flow fields $\mathbf{w}$ and occlusion masks $\mathbf{m}$ are estimated by this module. The Local Neural Texture Renderer renders the target images with vivid source features using the local attention blocks. 
We describe the details of these modules in the following subsections. 
Please note that to simplify the notations, we describe the network with a single local attention block. As shown in Figure~\ref{fig:OurModel}, our model can be extended to use multiple attention blocks at different scales.

\subsection{Global Flow Field Estimator}
\label{sec:global_flow}

We use the 18-channel heat map that encodes the locations of 18 joints of a human body as the structure guidance. Following the previous works~\cite{ma2017pose,siarohin2018deformable,zhu2019progressive}, the human body joints are detected by the Human Pose Estimator~\cite{cao2017realtime}.
Let $\mathbf{p}_s$ and $\mathbf{p}_t$ denote the structure guidance of the source image $\mathbf{x}_s$ and the target image $\mathbf{x}_t$ respectively. 
The Global Flow Field Estimator $F$ takes $\mathbf{x}_s$, $\mathbf{p}_s$, and $\mathbf{p}_t$ as inputs and generates the flow fields $\mathbf{w}$ and occlusion masks $\mathbf{m}$.

\begin{equation}
    \mathbf{w},\mathbf{m}=F(\mathbf{x}_s, \mathbf{p}_s, \mathbf{p}_t)
\end{equation}
where the flow fields $\mathbf{w}$ assign a source patch for each target location. The occlusion masks $\mathbf{m}$ with continuous values between $0$ and $1$ indicate whether the flowed source patches can be used to generate targets. We design $F$ as a fully convolutional network. $\mathbf{w}$ and $\mathbf{m}$ share all weights of $F$ other than their output layers.


Warping sources at the feature level can help models to be able to generate new content. Meanwhile, it relaxes the requirements of the flow field estimation since the resolutions of the generated flow fields are reduced.
However, networks may struggle to find reasonable sampling positions. An important reason is that the gradient propagation of the input features and flow fields are mutually constrained during the warping operation. The input features cannot obtain correct gradients without reasonable flow fields and vice versa. Therefore, we use additional losses to help with the training.
We propose a sampling correctness loss to constrain $\mathbf{w}$ in a self-supervised manner. The sampling correctness loss calculates the similarity between the warped source feature and the ground-truth target feature at the VGG feature level. 
Let $\mathbf{v}_{s}$ and $\mathbf{v}_{t}$ denote the features generated by a specific layer of VGG19. $\mathbf{v}_{s,\mathbf{w}}=\mathbf{w}(\mathbf{v}_{s})$ is the warped results of the source feature $\mathbf{v}_{s}$ using $\mathbf{w}$.
The sampling correctness loss calculates the relative cosine similarity between $\mathbf{v}_{s,\mathbf{w}}$ and $\mathbf{v}_{t}$.
\begin{equation}
   \mathcal{L}_{c} = \frac{1}{N}\sum_{l\in \Omega}exp(-\frac{\mu(\mathbf{v}_{s,\mathbf{w}}^l, \mathbf{v}_{t}^l)}{\mu_{max}^l})
\end{equation}
where $\mu(*)$ denotes the cosine similarity. Coordinate set $\Omega$ contains all $N$ positions in the feature maps.
$\mathbf{v}_{s,\mathbf{w}}^l$ and $\mathbf{v}_{t}^l$ denote the features of $\mathbf{v}_{s,\mathbf{w}}$ and $\mathbf{v}_{t}$ located at the coordinate $l=(x, y)$. 
The normalization term $\mu_{max}^l$ is used to avoid the bias brought by occlusion. It represents the similarity between $\mathbf{v}_t^l$ and its most similar feature in the source feature map $\mathbf{v}_s$. It is calculated as
\begin{equation}
   \mu_{max}^l = \max\limits_{l'\in \Omega}\mu(\mathbf{v}_{s}^{l'}, \mathbf{v}_{t}^{l})
\end{equation}
where $\mathbf{v}_{s}^{l'}$ is the feature of $\mathbf{v}_s$ located at the coordinate $l'$.

\begin{figure*}[t]
\begin{center}
\includegraphics[width=0.95\linewidth]{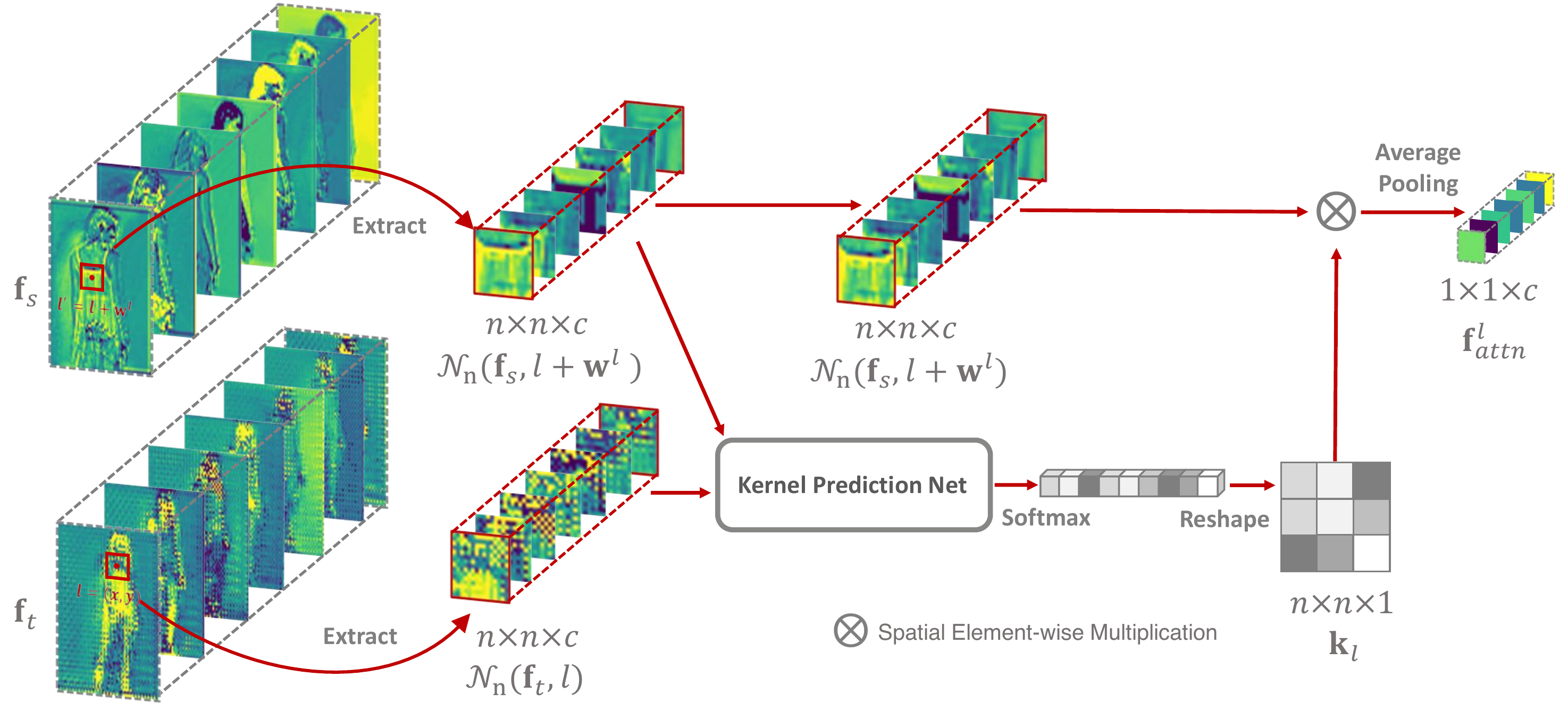}
\end{center}
   \caption{Overview of our Local Attention. We first extract the feature patch pair from the source and target according to the flow fields. Then the context-aware sampling kernel is calculated by the kernel prediction net. Finally, we sample the source feature and obtain the warped result located at $l$.}
 
\label{fig:LocalAttention}
\end{figure*}

Our sampling correctness loss calculates the element-wise similarities. It cannot model the correlation of adjacent features. However, the deformations of image neighborhoods are highly correlated. 
To model these correlations, we further propose a regularization term.
This regularization term is designed to punish local regions where the transformation is not an affine transformation. Let $\mathbf{c}_{t}$ be the 2D coordinate matrix of the target feature map. The corresponding source coordinate matrix can be written as $\mathbf{c}_{s} = \mathbf{c}_{t}+\mathbf{w}$. 
We use $\mathcal{N}_n(\mathbf{c}_{t}, l)$ to denote local $n\times n$ patch of $\mathbf{c}_{t}$ centered at the location $l$.
Our regularization assumes that the transformation between $\mathcal{N}_n(\mathbf{c}_{t}, l)$ and $\mathcal{N}_n(\mathbf{c}_{s}, l)$ is an affine transformation.
\begin{equation}
    \mathbf{T}_l={\mathbf{A}}_l\mathbf{S}_l=\left[ {\begin{array}{ccc}\theta_{11}&\theta_{12}&\theta_{13}\\
    \theta_{21}&\theta_{22}&\theta_{23}\end{array}} \right]\mathbf{S}_l
\end{equation}
where $\mathbf{T}_l=\left[{\begin{array}{cccc}x_1&x_2&...&x_{n\times n}\\
    y_1&y_2&...&y_{n \times n}\end{array}}\right]$ with each coordinate $(x_i,y_i)\in \mathcal{N}_n(\mathbf{c}_t, l)$ and $\mathbf{S}_l=\left[{\begin{array}{cccc}x_1&x_2&...&x_{n\times n}\\
    y_1&y_2&...&y_{n\times n}\\1&1&...&1\end{array}}\right]$ with each coordinate $(x_i,y_i)\in \mathcal{N}_n(\mathbf{c}_s, l)$.
    The estimated affine transformation parameters $\hat{\mathbf{A}}_l$ can be solved using the least-squares estimation as  
\begin{equation}
    \hat{\mathbf{A}}_l = \mathbf{T}_l\mathbf{S}_l^H(\mathbf{S}_l\mathbf{S}_l^H)^{-1}
\end{equation}
where $\mathbf{S}_l^H$ is the transpose matrix of $\mathbf{S}_l$.
Our regularization is calculated as the $\ell_2$ distance of the error.
\begin{equation}
    \mathcal{L}_r=\sum_{l \in \Omega} \left\lVert \mathbf{T}_l - \hat{\mathbf{A}}_l{\mathbf{S}}_l \right\lVert_2^2
\end{equation}


\subsection{Local Neural Texture Renderer}
\label{Local_attention}
The Local Neural Texture Renderer $G$ is responsible for generating the result images by rendering target poses with the source neural textures.
It takes $\mathbf{x}_s$, $\mathbf{p}_t$, $\mathbf{w}$, and $\mathbf{m}$ as inputs and generates the result image $\hat{\mathbf{x}}_t$.
\begin{equation}
    \hat{\mathbf{x}}_t=G(\mathbf{x}_s, \mathbf{p}_t, \mathbf{w},\mathbf{m})
\end{equation}

To avoid the poor gradient propagation of the Bilinear sampling, we propose a local attention operation to sample the source features with a content-aware manner. 
Our local attention works as a neural renderer where the source neural textures are sampled to render the target bones.
We illustrate the processing details in Figure~\ref{fig:LocalAttention}.
Let $\mathbf{f}_t$ and $\mathbf{f}_s$ represent the extracted features of target bones $\mathbf{p}_t$ and source images $\mathbf{x}_s$ respectively. 
For each location $l$, local patches $\mathcal{N}_n(\mathbf{f}_t, l)$ and $\mathcal{N}_n(\mathbf{f}_s, l+\mathbf{w}^l)$ are first extracted from $\mathbf{f}_t$ and $\mathbf{f}_s$.\footnote{The patch $\mathcal{N}_n(\mathbf{f}_s, l+\mathbf{w}^l)$ is extracted using the Bilinear sampling as the coordinates may not be integers.}
Then, we predict the local $n \times n$ kernel $\mathbf{k}_l$ as the attention coefficients from the extracted local feature patch pair using a kernel prediction network $M$. 
\begin{equation}
    \mathbf{k}_l=M(\mathcal{N}_n(\mathbf{f}_s, l+\mathbf{w}^l), \mathcal{N}_n(\mathbf{f}_t, l))
\end{equation}
We design $M$ as a fully connected network.
The local patch pair $\mathcal{N}_n(\mathbf{f}_s, l+\mathbf{w}^l)$ and $\mathcal{N}_n(\mathbf{f}_t, l)$ are directly concatenated as the network inputs.
We use the softmax function as the non-linear activation function of the output layer of model $M$.
This operation forces the sum of $\mathbf{k}_l$ to 1, which enables the stability of gradient backward. Finally, the attention result localed at coordinate $l=(x,y)$ is calculated as 
\begin{equation}
    \mathbf{f}^l_{attn}=P(\mathbf{k}_l\otimes\mathcal{N}_n(\mathbf{f}_s, l+\mathbf{w}^l))
\end{equation}
where $\otimes$ denotes the element-wise multiplication over the spatial domain and $P$ represents the global average pooling operation. 
The final warped feature $\mathbf{f}_{attn}$ is obtained by repeating the previous steps for each location $l$. 

Furthermore, in order to enable the network to generate occluded contents, we use the mask $\mathbf{m}$ with continuous values between 0 and 1 to select features between the warped result $\mathbf{f}_{attn}$ and the target feature $\mathbf{f}_{t}$. 
The final output feature map $\mathbf{f}_{out}$ is calculated as 
\begin{equation}
    \mathbf{f}_{out} = (\mathbf{1}-\mathbf{m})*\mathbf{f}_t + \mathbf{m}*\mathbf{f}_{attn}
\end{equation}

We train the network using a joint loss consisting of a reconstruction $\ell_1$ loss, adversarial loss, perceptual loss, and style loss.
The reconstruction $\ell_1$ loss is written as 
\begin{equation}
    \mathcal{L}_{\ell_1}=\left\lVert \mathbf{x}_t - \hat{\mathbf{x}}_t \right\lVert_1
\end{equation}
The generative adversarial loss~\cite{goodfellow2014generative} is used to mimic the distributions of the ground-truth $\mathbf{x}_{t}$.
\begin{align}
   \mathcal{L}_{adv} &= \mathbb{E}[\log(1-D(G(\mathbf{x}_s, \mathbf{p}_t, \mathbf{w},\mathbf{m})))] \notag \\
   &+ \mathbb{E}[\log D(\mathbf{x}_{t})]
   \label{eq:advr}   
\end{align}
where $D$ is the discriminator of the Local Neural Texture Renderer $G$. 
The perceptual loss and style loss introduced by~\cite{johnson2016perceptual} are used to reduce the reconstruction errors. The perceptual loss calculates $\ell_1$ distance between activation maps of a pre-trained network. It can be written as
\begin{equation}
    \mathcal{L}_{perc}=\sum_i\left\lVert \phi_i(\mathbf{x}_t) - \phi_i(\hat{\mathbf{x}}_t)\right\lVert_1
\end{equation}
where $\phi_i$ is the activation map of the $i$-th layer of a pre-trained network. The style loss calculates the statistic error between the  activation maps as
\begin{equation}
    \mathcal{L}_{style}=\sum_j\left\lVert G_j^\phi(\mathbf{x}_t) - G^\phi_j(\hat{\mathbf{x}}_t)\right\lVert_1
\end{equation}
where $G_j^\phi$ is the Gram matrix constructed from activation maps $\phi_j$.
Our GFLA model is trained using the overall loss as
\begin{align}
   \mathcal{L}_G &= \lambda_c\mathcal{L}_{c} + \lambda_r\mathcal{L}_{r} + \lambda_{\ell_1}\mathcal{L}_{\ell_1}+ \lambda_a\mathcal{L}_{adv} \notag \\
   &+ \lambda_p\mathcal{L}_{prec}+ \lambda_s\mathcal{L}_{style}
   \label{eq:joint_loss}   
\end{align}

\section{Modeling The Temporal Consistency \\ for Person Image Animation}

The pose-guided person image animation task refers to generating videos by rendering continuous skeletons using the neural textures of source images.
Different from the generation task, it requires not only generating realistic textures for each frame but also modeling the temporal consistency between adjacent frames.
Therefore, we further improve our model to generate coherent results. First, a Motion Extraction Network is proposed to extract accurate movements from the noisy input skeletons. Then we improve our GFLA model to generate sequences in a recurrent manner. We describe the details in this section.

\subsection{Motion Extraction Network}
\label{sec:MEN}
\begin{figure}[t]
\begin{center}
\includegraphics[width=1\linewidth]{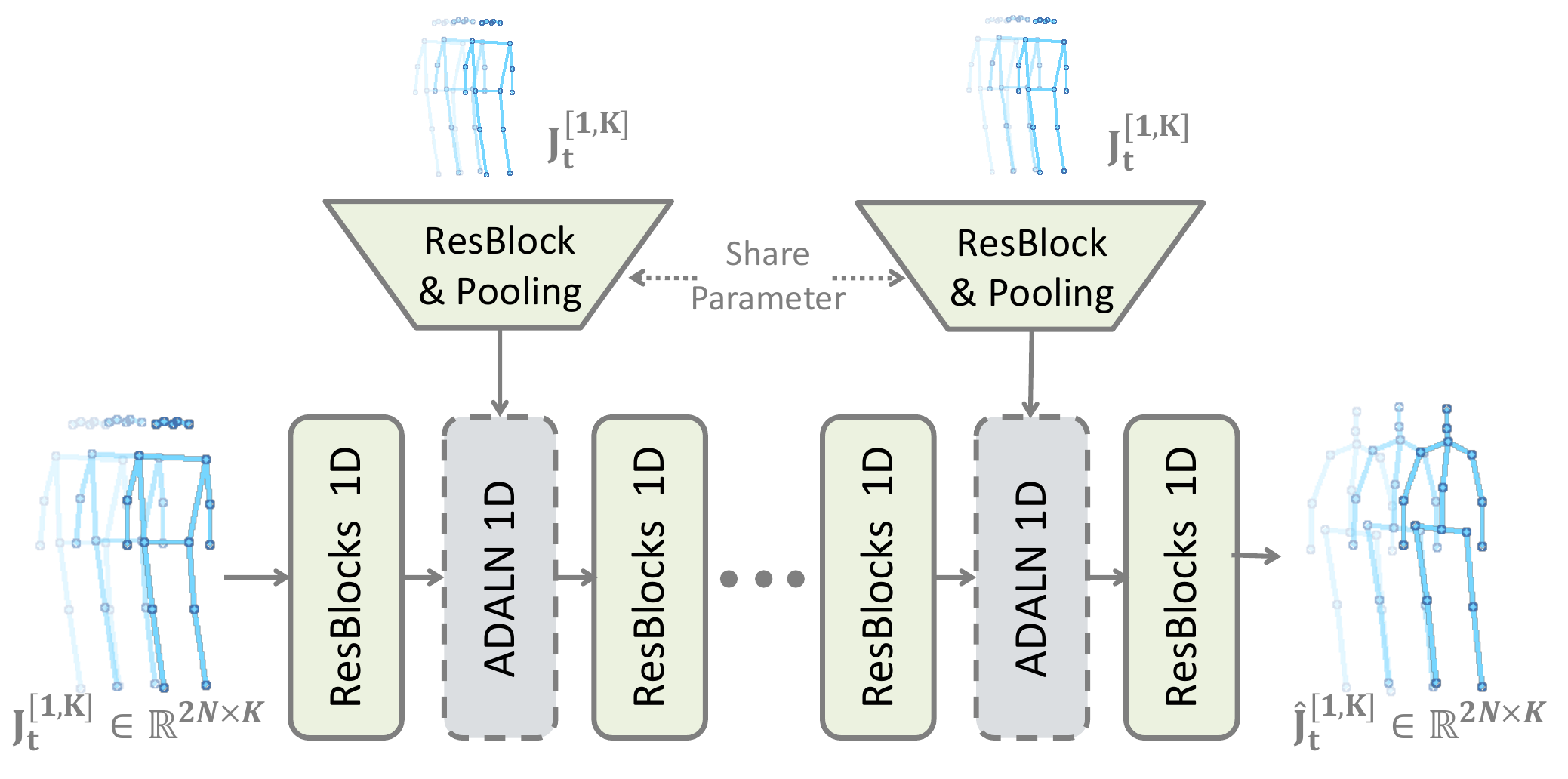}
\end{center}
   \caption{The architecture of our Motion Extraction Network.}
 
\label{fig:MotionExtractNet}
\end{figure}

One of the major problems is that the input skeleton sequences extracted by the popular algorithms~\cite{cao2018openpose,li2018crowdpose} are not temporally consistent. As shown in Figure~\ref{fig:ablation_video}, the predicted locations vibrate around the ground-truth values. Our Motion Extraction Network works as a denoise model extracting accurate movements from noisy skeleton sequences. The architecture of the Motion Extraction Network is shown in Figure~\ref{fig:MotionExtractNet}. Inspired by the paper~\cite{pavllo20193d}, we design the network using 1D convolutional layers.
The input layer of this network takes the concatenated $(x, y)$ coordinates of the $N$ joints for each skeleton frame instead of the 2D heat maps. Let $\mathbf{J}_t^{[1,K]} \in \mathbb{R}^{2N \times K}$ denotes the joints of $K$ input skeletons. 
The output joints $\mathbf{\hat{J}}_t^{[1,K]}$ contains the coordinates of skeletons with accurate movements. 
We use Adaptive layer normalization (ADALN) in this network. It has a similar architecture to that of ANAIN~\cite{huang2017arbitrary} but using layer normalization~\cite{ba2016layer} as the normalization function. Layer normalization calculates the statistics for each single training case and normalizes the activities in a batch-wise manner. The effect of this normalization operation can be explained as to removing the unrelated factors such as global locations and scales, thereby making the network focus on motion extraction. As our task is to reconstruct the coherent skeletons, we need to recover the statistics of the original sequences after reasoning about the motions. Therefore, we enable the network to recover the original statistics by calculating the affine parameters of the normalization layers from the input skeletons. 
The network is trained with ground-truth joints $\mathbf{J}_{gt}^{[1,K]}$. The commonly used mean per-joint position error (MPJPE) is employed as the loss function.
\begin{equation}
\label{eq:MPJPE}
  \mathcal{L}_{mpjpe}= \left\lVert \mathbf{\hat{J}}_t^{[1,K]}, \mathbf{J}_{gt}^{[1,K]},  \right\lVert_1
\end{equation}

Since most person animation datasets do not provide the required ground-truth skeleton labels, we train this network separately using the Human3.6M dataset~\cite{ionescu2013human3}. This dataset contains accurate 3D human skeleton sequences acquired by recording the performance of 11 subjects under 4 different viewpoints. We extract the noise skeleton inputs from the videos of the Human3.6M dataset by using the pose extractor~\cite{li2018crowdpose}. The ground-truth labels $\mathbf{J}_{gt}^{[1,K]}$ are obtained by projecting the 3D skeletons to the corresponding viewpoints. After training the Motion Extraction Network, we can obtain the clean skeletons $\hat{\mathbf{J}}_{t}^{[1,K]}$ by performing inference on the animation datasets.

\subsection{Sequential Global-Flow Local-Attention Model}
\label{sequential_GFLA}
\begin{figure}[t]
\begin{center}
\includegraphics[width=1\linewidth]{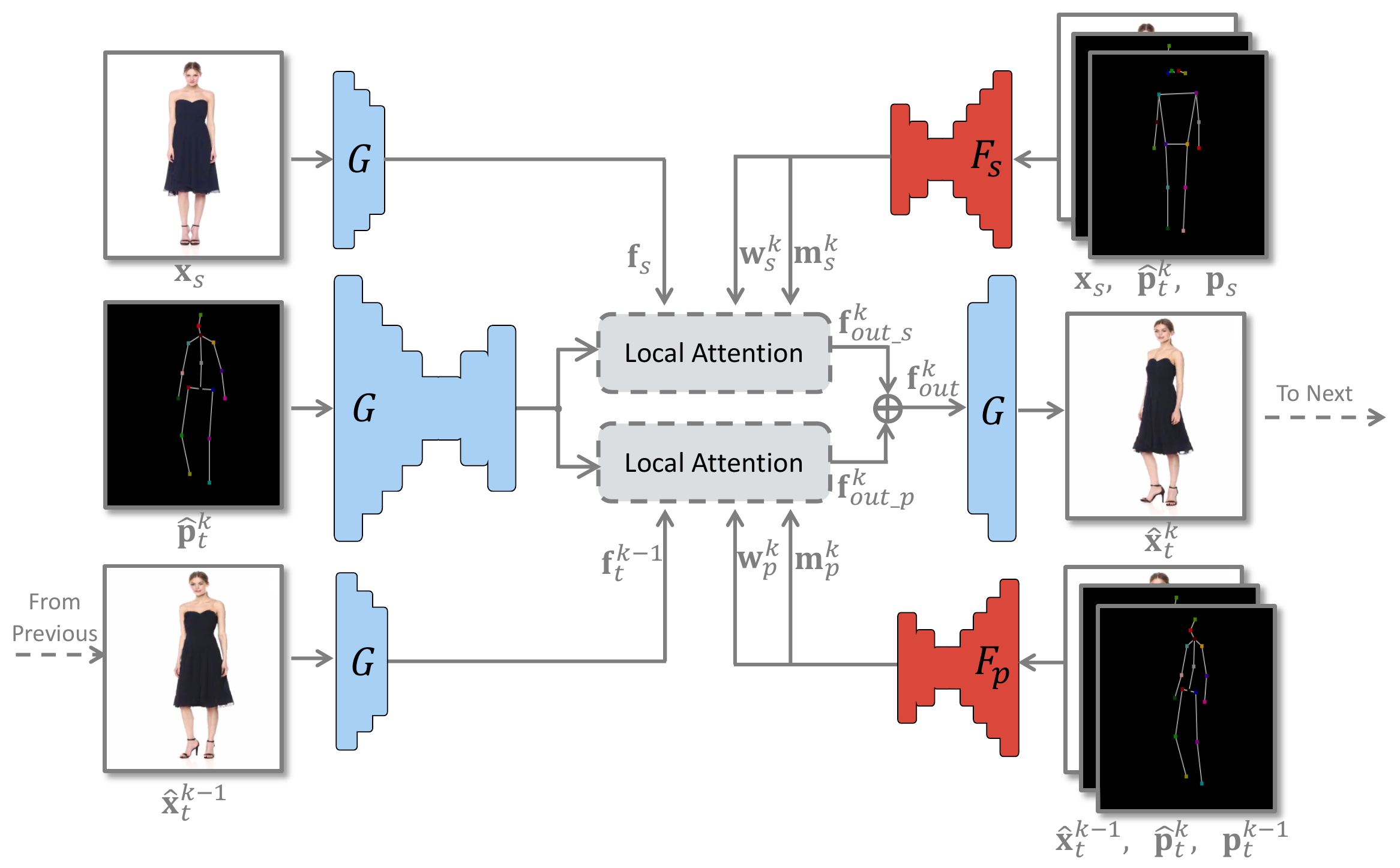}
\end{center}
   \caption{The generation process of video frame $\mathbf{\hat{x}}_t^k$. Our sequential GFLA model spatially transforms the information of the source image $\mathbf{x}_s$ and previously generated image $\mathbf{\hat{x}}_t^{k-1}$ to generate the result.}
 
\label{fig:recurrentGFLA}
\end{figure}

We design a sequential GFLA model to generate result videos from the extracted accurate movements. 
Let $\mathbf{\hat{p}}^{[1,K]}_t \equiv \{\mathbf{\hat{p}}_t^1,\mathbf{\hat{p}}_t^2,...,\mathbf{\hat{p}}_t^K \}$ denotes the 2D heat map sequences obtained from the extracted joints $\mathbf{\hat{J}}_t^{[1,K]}$. Our model generates video clips $\mathbf{\hat{x}}_t^{[1,K]} \equiv \{\mathbf{\hat{x}}_t^1,\mathbf{\hat{x}}_t^2,...,\mathbf{\hat{x}}_t^K \}$ by rendering skeletons $\mathbf{\hat{p}}^{[1,K]}_t$ using the appearance of the source image $\mathbf{x}_s$. 
We explicitly build the correlations between adjacent frames. Video clips are generated in a recurrent manner: the previously generated frames are used as the inputs of the current generation step.
Specifically, Figure~\ref{fig:recurrentGFLA} shows the generation process of frame $\mathbf{\hat{x}}_t^k$.
It can be seen that we have added an additional spatial transformation module responsible for transforming the information of the previously generated frame $\mathbf{\hat{x}}_t^{k-1}$ to the sequential GFLA model.
Our model first extracts flow fields $\mathbf{w}_s^k$ and $\mathbf{w}_p^k$ using the Global Flow Field Estimators $F_s$ and $F_p$ respectively.

\begin{align}
  &\mathbf{w}_s^k,\mathbf{m}_s^k =F_s(\mathbf{x}_s, \mathbf{p}_s, \mathbf{\hat{p}}_t^k)   \\
  &\mathbf{w}_p^k,\mathbf{m}_p^k =F_p(\mathbf{\hat{x}}_t^{k-1}, \mathbf{\hat{p}}_t^{k-1}, \mathbf{\hat{p}}_t^k)     
\end{align}
where the $\mathbf{m}_s^k$ and $\mathbf{m}_p^k$ are the occlusion masks. 
The Local Neural Texture Renderer $G$ is then used to generate the result image by spatially transforming the information of $\mathbf{x}_s$ and $\mathbf{\hat{x}}_t^{k-1}$.
\begin{equation}
  \mathbf{\hat{x}}_t^k = G(\mathbf{x}_s, \mathbf{p}_s,  \mathbf{w}_s^k,\mathbf{m}_s^k, \mathbf{\hat{x}}_t^{k-1}, \mathbf{\hat{p}}_t^{k-1}, \mathbf{w}_p^k,\mathbf{m}_p^k, \mathbf{\hat{p}}_t^{k})
\end{equation}
Two local attention modules are used to warp the features of the source image $\mathbf{x}_s$ and previously generated image $\mathbf{\hat{x}}_t^{k-1}$. 
The processing operation is the same as that described in Section~\ref{Local_attention}. The output features $\mathbf{f}_{{out}\_s}^k$ and $\mathbf{f}_{{out}\_p}^k$ are generated by these local attention modules.
The final output feature $\mathbf{f}_{out}^k$ is calculated by fusing the outputs of the two branches 
\begin{equation}
  \mathbf{f}_{out}^k = \mathbf{f}_{out\_s}^k + \mathbf{f}_{out\_p}^k
\end{equation}

We train the animation model using both spatial and temporal losses.
The spatial losses can constrain the model to generate realistic frames. We use the same joint loss (Equation~\ref{eq:joint_loss}) as that of our image generation model for each result frame. The temporal loss is used to model the correlations between different frames. We use a temporal discriminator $D_{v}$ to calculate this loss. The temporal discriminator $D_{v}$ takes image sequences as inputs and estimates the probabilities that the inputs are sampled from real video clips. 
\begin{align}
   \mathcal{L}_{adv\_v} &= \mathbb{E}[\log(1-D_v(\mathbf{\hat{x}}_t^{[1,K]})))] \notag \\
   &+ \mathbb{E}[\log D_v(\mathbf{x}_{t}^{[1,K]})]
   \label{eq:Tempal}   
\end{align}
Therefore, the overall loss function of our animation model can be written as 
\begin{equation}
  \mathcal{L}_{A} = \frac{1}{K}\sum_{k=1}^K \mathcal{L}_G^k + \lambda_{v}\mathcal{L}_{adv\_v}
\end{equation}
where $\mathcal{L}_G^k$ represents the spatial loss of frame $\mathbf{\hat{x}}_t^k$. 







%


\section{Experiments}
In this section, we evaluate the performance of the proposed method. The network structures and training details are first provided in Section~\ref{implementation_details}. Then, we verify the impact of the proposed modules. The ablation studies are divided into two parts. In Section~\ref{spatial_improvements}, we show that our GFLA framework can efficiently spatially transform source feature maps. In Section~\ref{temporal_improvements}, we verify the efficacy of our sequential GFLA model. Finally, we compare our method with several state-of-the-art algorithms over both generation and animation tasks in Section~\ref{comparisons}.

\subsection{Implementation Details} 
\label{implementation_details}
\noindent
\textbf{Network Architecture and Training Details.}
Auto-encoder structures are employed to design our networks. We use the residual block~\cite{he2016deep} as the basic component of our model. Unless otherwise specified, we train our models using $256 \times 256$ images. We use local attention modules for feature maps with resolutions of $32 \times 32$ and $64 \times 64$. The extracted local patch sizes are $3$ and $5$ respectively. 
For the person image generation task, we train our GFLA model in stages. The Flow Field Estimator is first trained to generate flow fields. Then we train the whole model in an end-to-end manner. For the image animation task, we first train the Motion Extraction Network using the Human3.6M dataset~\cite{ionescu2013human3} as described in Section~\ref{sec:MEN}.
Then we train our sequential GFLA model using the predicted clean skeletons. We adopt the ADAM optimizer with the learning rate as $10^{-4}$.

\noindent
\textbf{Metrics.} We employ both image-based metrics and video-based metrics to evaluate our results. Learned Perceptual Image Patch Similarity~\cite{zhang2018unreasonable} (LPIPS) is used to calculate the reconstruction errors of generated images. This metric computes perceptual distances between input image pairs. Meanwhile, we employ Fr\'echet Inception Distance~\cite{heusel2017gans} (FID) to measure the realism of the generated images. It calculates the Wasserstein-2 distance between distributions of the generated data and real data. For video results, in order to model the temporal consistency errors, we use the I3D model~\cite{carreira2017quo} to extract the video features. 
Average Euclidean Distance~\cite{Siarohin_2019_CVPR} (AED) is used as the perceptual reconstruction error indicator. It calculates the Euclidean distance between features of generated videos and ground-truth videos. FID-Video takes the extracted video features as inputs and evaluates the realism of generated videos.
Besides, we perform a Just Noticeable Difference (JND) test to evaluate the subjective quality. Volunteers are asked to choose the more realistic image from the data pair of ground-truth and generated images. We provide the fooling rate as the evaluation result.





\noindent
\textbf{Datasets.} 
For the person image generation task, we use two public datasets: Market-1501~\cite{zheng2015scalable} and DeepFashion In-shop Clothes Retrieval Benchmark~\cite{liu2016deepfashion}. Market-1501 contains 32668 low-resolution images (128 $\times$ 64). The images vary in terms of the viewpoints, background, illumination, etc. The DeepFashion dataset contains 52712 high-quality model images with clean backgrounds. We split these datasets with the same method as that of~\cite{zhu2019progressive}. The personal identities of the training and testing sets do not overlap. Two video datasets are used for animation tasks: FashionVideo~\cite{zablotskaia2019dwnet} and iPER~\cite{liu2019liquid}. The FashionVideo dataset contains 500 training and 100 test videos, each containing roughly 350 frames.  Videos have static viewpoints and clean backgrounds. The iPER dataset contains 206 high-resolution videos. Human subjects in this dataset have different conditions of shape, height, and gender. 










\subsection{Efficacy of the GFLA Framework}
\label{spatial_improvements}
We evaluate the components of our GFLA framework by comparing our model with the following variants. 

\noindent
\textbf{Baseline.}
An auto-encoder convolutional network is used as the Baseline model. We do not use any attention blocks in this model. Source images $\mathbf{x}_s$ and guidance poses $\mathbf{p}_t$, $\mathbf{p}_s$ are directly concatenated as the model inputs.

\noindent
\textbf{Global Attention Model (Global-Attn).}
The Global-Attn model is used to compare global attention with our local attention. This model has a similar architecture to the Local Neural Texture Renderer $G$ in Section~\ref{Local_attention}. The local attention modules are replaced by global attention blocks where the attention coefficients are calculated by the similarities between the source features $\mathbf{f}_s$ and target features $\mathbf{f}_t$.

\noindent
\textbf{Local Attention Model (Local-Attn).}
The Local-Attn model is used to evaluate the efficacy of our sampling correctness loss and regularization loss described in Section~\ref{sec:global_flow}. We use the same network architecture as the GFLA model. 
However, the sampling correctness loss $\mathcal{L}_c$ and regularization loss $\mathcal{L}_r$ are not employed for training.



\noindent
\textbf{Bilinear Sampling Model (Bi-Sample).}
The Bi-Sample model is used to evaluate the efficacy of our local-attention module described in Section~\ref{Local_attention}. We use both Global Flow Field Estimator $F$ and Local Neural Texture Renderer $G$ in this model. However, the local-attention module is replaced by the Bilinear sampling method.



\noindent
\textbf{Full Generation Model (Ours).}
The proposed GFLA framework is used in this model. 


\begin{table}[t]
\centering
\setlength\extrarowheight{1pt}
\resizebox{.5\textwidth}{!}{%
\centering
\begin{tabular}{c||c|c|P{1.5cm}|P{1.5cm}}
\hline
&Flow-Based&Content-aware & \multirow{2}{*}{FID}& \multirow{2}{*}{LPIPS} \\ 
&Method&Sampling&&  \\\hline
&&&&\\ [-8pt]\hline
Baseline     & N          & -                      & 16.008 & 0.2473 \\
Global-Attn  & N          & -                      & 18.616 & 0.2575 \\
Local-Attn   & Y          & Y                      & 12.943 & \textbf{0.2339} \\
Bi-Sample    & Y          & N                      & 12.143 & 0.2406 \\
Full Model   & Y          & Y                      & \textbf{10.573} & 0.2341 \\ \hline
\end{tabular}%
}
\caption{The ablation study results of our GFLA model.}
\label{tab:AblationStudy}
\end{table}

\begin{table*}
\setlength\extrarowheight{1pt}
\centering
\resizebox{.9\textwidth}{!}{%

\begin{tabular}{p{2cm}||P{1.2cm}P{1.2cm}P{1.2cm}||P{1.2cm}P{1.2cm}P{2cm}P{1.2cm}||P{2cm}}
\hline
          & \multicolumn{3}{c||}{DeepFashion} & \multicolumn{4}{c||}{Market-1501}&  Number of \\ \cline{2-8} 
          & FID        & LPIPS      & JND    & FID    & LPIPS  & Mask-LPIPS & JND & Parameters\\ \hline
          &            &            &        &        &        &            & & \\ [-8pt]\hline
Def-GAN   & 18.457     & 0.2330     & 9.12\%     & 25.364 & 0.2994 & 0.1496     &23.33\%     & 82.08M\\ \hline
VU-Net    & 23.667     & 0.2637     & 2.96\% & 20.144 & 0.3211 & 0.1747     & 24.48\% & 139.36M\\ \hline
Pose-Attn & 20.739     & 0.2533     & 6.11\% & 22.657 & 0.3196 & 0.1590     & 16.56\%  & 41.36M\\ \hline
Intr-Flow & 16.314     & \textbf{0.2131} & 12.61\%  & 27.163 & 0.2888 & \textbf{0.1403}     & \textbf{30.85}\%      &    49.58M      \\ \hline
Ours      & \textbf{10.573}& 0.2341 &\textbf{24.80}\%      & \textbf{19.751} & \textbf{0.2817} & 0.1482&27.81\%    & \textbf{14.04M}\\ \hline
\end{tabular}%

}

\caption{Quantitative comparisons over dataset DeepFashion~\cite{liu2016deepfashion} and Market-1501~\cite{zheng2015scalable} with state-of-the-art person image generation methods including Def-GAN~\cite{siarohin2018deformable}, VU-Net~\cite{esser2018variational}, Pose-Attn~\cite{zhu2019progressive}, and Intr-Flow~\cite{li2019dense}. FID~\cite{heusel2017gans} and LPIPS~\cite{zhang2018unreasonable} are objective metrics. JND is obtained by human subjective studies.}

\label{tab:object}
\end{table*}
The evaluation results of the ablation study are shown in Table~\ref{tab:AblationStudy}. Compared with the Baseline, the performance of the Global-Attn model is degraded.
It means that the global attention cannot efficiently transform the source information in this task.
Flow-based models (Local-Attn, Bi-Sample, and our Full model) improve the generation results, since they force the attention coefficient matrix to be a sparse matrix.
The Local-Attn model achieves a good LPIPS result. However, the poor FID score indicates that the realism of its results is degraded since it cannot find reasonable sampling positions for target outputs.
The Bi-Sample model is able to obtain relatively accurate flow fields. 
However, the pre-defined sampling method with limited receptive fields leads to performance degradation.
Our full model improves the performance by using the content-aware sampling operation with adjustable receptive fields.


\begin{figure}[t]
\begin{center}
\includegraphics[width=1\linewidth]{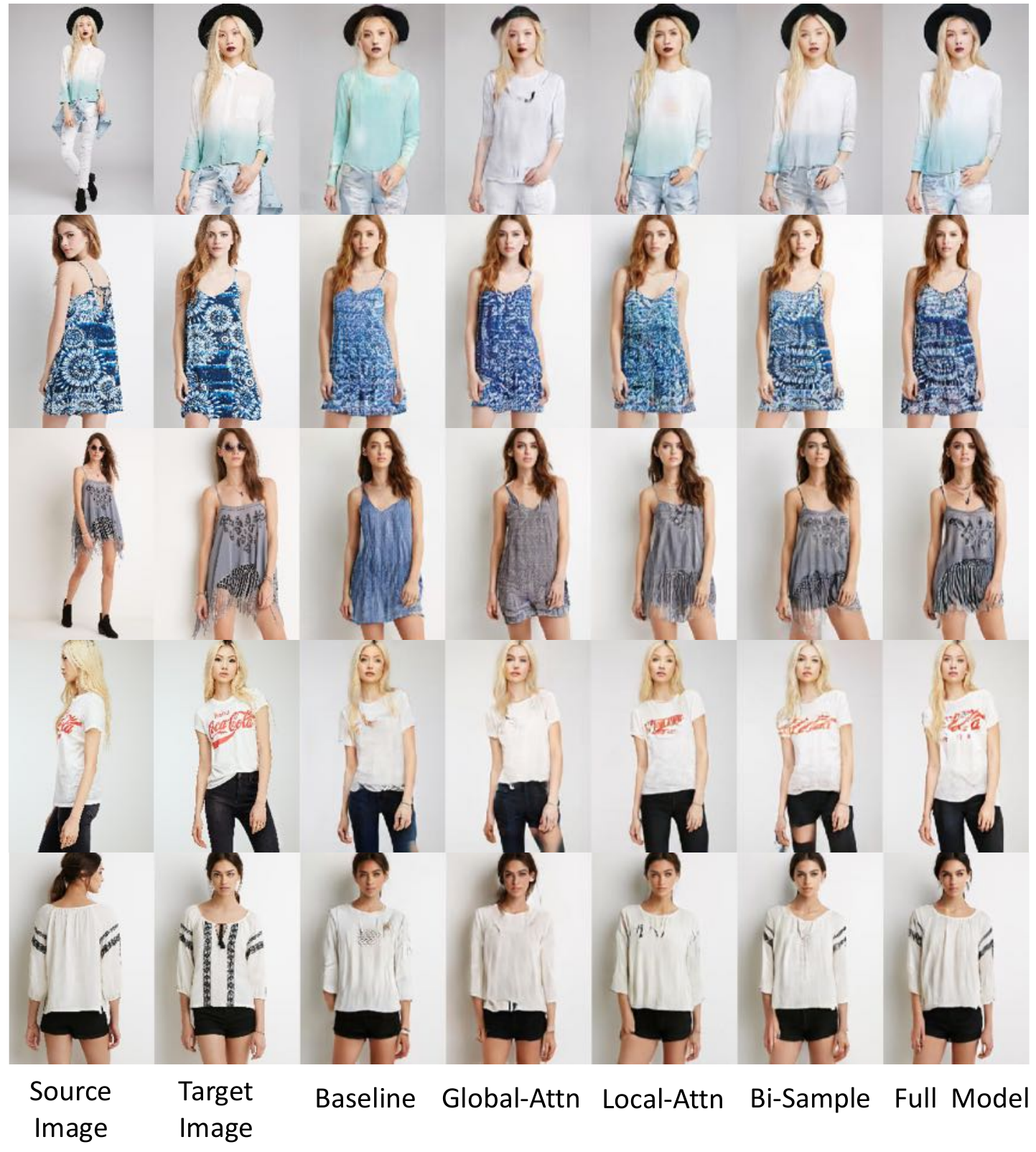}
\end{center}
   \caption{Qualitative results of our GFLA model and its variants.}
 
\label{fig:ablation}
\end{figure}

\begin{figure}[t]
\begin{center}
\includegraphics[width=1\linewidth]{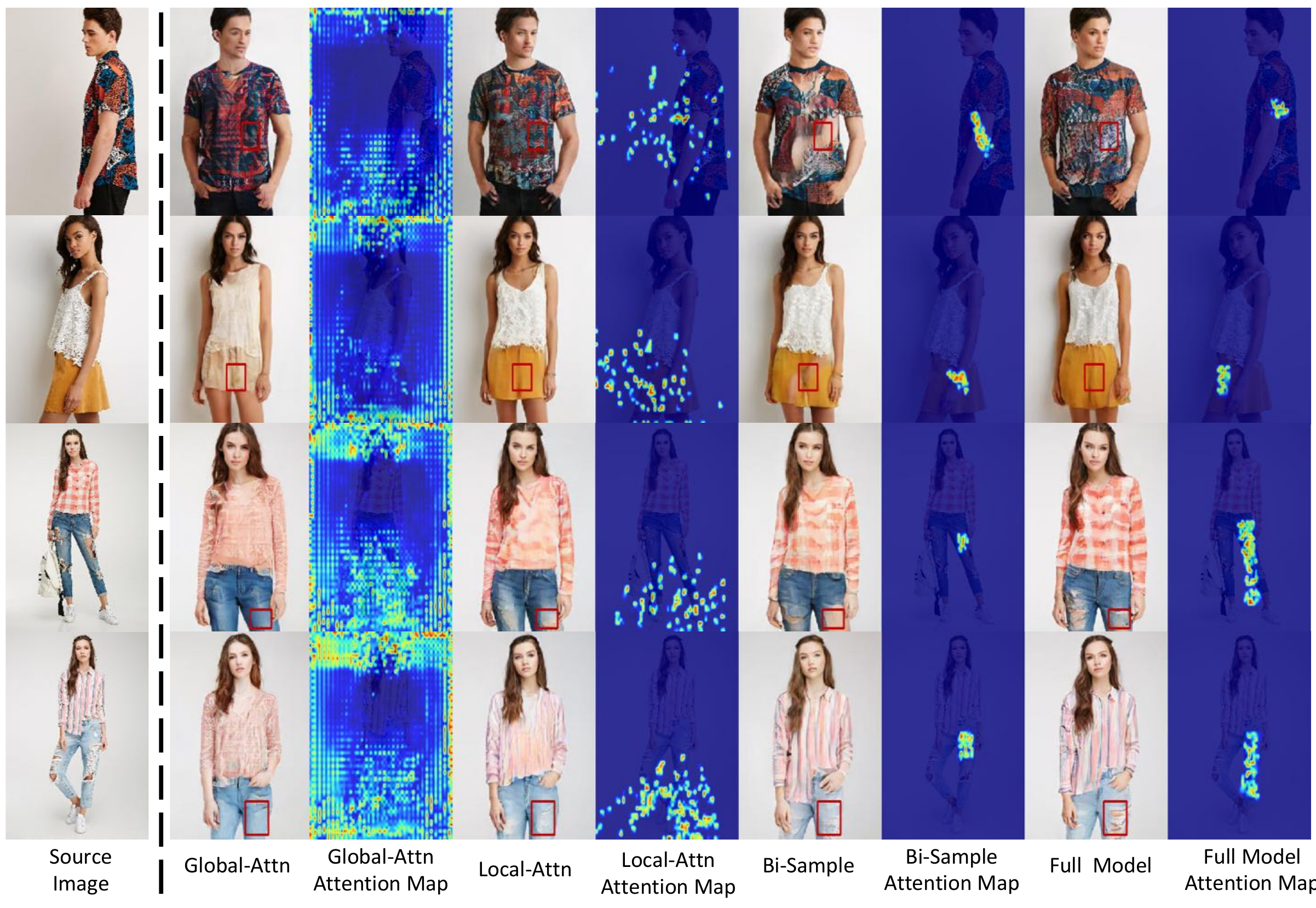}
\end{center}
   \caption{The visualization results of different attention modules. The red rectangles in the generated images indicate the query locations. The heat maps show the visualization of the corresponding attention coefficients of these query locations. Blue represents low weights.}
 
\label{fig:ablation_visi_attn}
\end{figure}

The subjective comparison of these ablation models can be found in Figure~\ref{fig:ablation}. The Baseline and Global-Attn model are able to generate images with correct poses. However, the source appearances are not well-maintained.
The possible explanation is that these models generate images by first extracting global features and then propagating the information to specific locations. 
However, the global features only characterize the global style of the sources, regardless of spatial information. Thus, it causes the vivid local texture details "wash away" in the ultimate image.
The flow-based methods spatially transform the features. They are able to reconstruct image details. The Local-Attn model generates textures with similar styles of that of sources. However, specific texture patterns (\emph{e.g.} logos) are not reconstructed, since it cannot extract accurate movements between sources and targets. The Bi-Sample model can generate vivid textures. However, artifacts can be observed in its results.
Our full model is able to generate realistic images. We further provide the visualization of the attention coefficients in Figure~\ref{fig:ablation_visi_attn}. 
For each attention module, we provide the generated target images and the corresponding attention coefficient heat maps. In order to visualize the attention coefficients, we first calculate attention maps of all query locations in the red rectangle. Then the visualization heat map is calculated by summing the obtained attention maps. 
It can be seen that the Global-Attn model struggles to exclude irrelevant information. Therefore, it is hard to reconstruct accurate textures using the attention results. 
The Local-Attn model casts a wide net to sample textures for a local target patch. It seems that this model tries to sample all possible positions and omitted the irrelevant information using occlusion masks. However, texture patterns are destroyed during this operation.
The Bi-Sample model is able to sample local regions. However, incorrect regions are often sampled due to the poor gradient propagation.
Our Full model using the content-aware sampling method can flexibly change the sampling weights and avoid artifacts.

\subsection{Efficacy of the temporal-consistency modeling}
\label{temporal_improvements}
We prove that our sequential GFLA model and Motion Extraction Network can help with modeling the temporal-consistency. We compare our model with the following variants.

\noindent
\textbf{Naive Animation Model (Naive-Animation).} We directly use our GFLA model described in Section~\ref{image_generation} as the Naive-Animation model. It is trained to generate a single target image from a source input. In the inference phase, video frames are generated independently.


\noindent
\textbf{Sequential Animation Model (Seq-Animation).} The Seq-Animation model is used to evaluate the efficacy of our sequential GFLA model. We use the architecture described in Section~\ref{sequential_GFLA} for this model. However, the Motion Extraction Network is not used to preprocess the noisy skeleton sequences.

\noindent
\textbf{Full Animation Model (Ours-Animation).} We use the sequential GFLA model with clean skeleton sequences obtained by the Motion Extraction Network.

\begin{figure}[t]
\begin{center}
\href{https://renyurui.github.io/GFLA-web/Animation_Ablation}{
\includegraphics[width=1\linewidth]{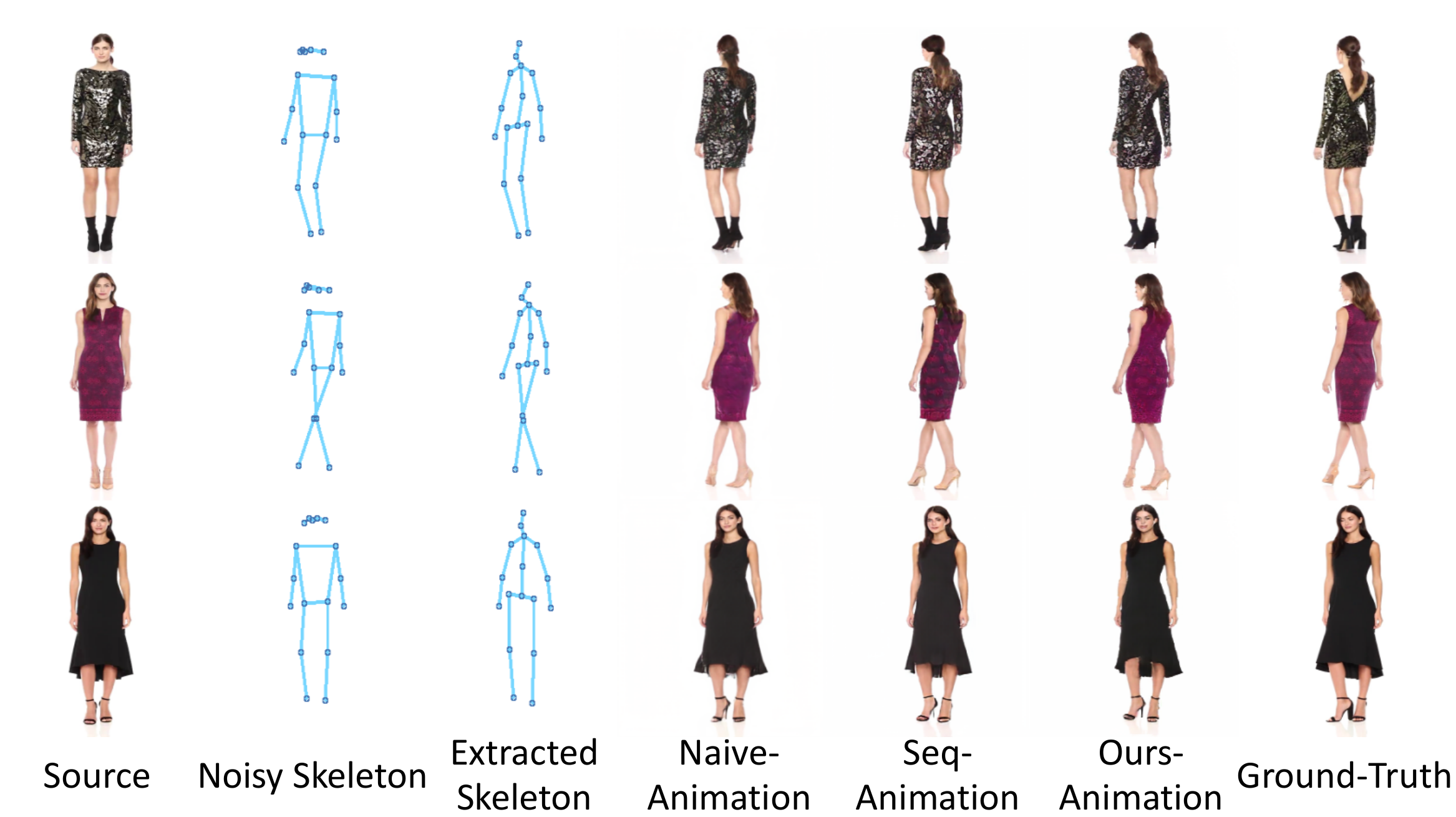}}
\end{center}
   \caption{Qualitative results of our person image animation model and its variants. \textit{Click on the image to play the video in a browser.} The red rectangles in the video highlight the temporal inconsistent clips.}
 
\label{fig:ablation_video}
\end{figure}

The evaluation results are shown in Table~\ref{tab:AblationStudyVideo}. It can be seen that obvious performance gain is obtained by using the sequential GFLA model which explicitly models the correlations between adjacent frames. However, the noisy input skeletons still cause inconsistency which leads to performance degradation. Our full model further improves the performance by extracting clean movements from the noisy input sequences.
The subjective results are shown in Figure~\ref{fig:ablation_video}. It can be seen that Naive-Animation can generate realistic video frames. However, it struggles to maintain temporal consistency.
The Seq-Animation model solves this problem to a certain extent. However, the noisy input sequences still cause incoherent results. By preprocessing the input skeletons using the Motion Extraction Network, our animation model is able to generate coherent videos with vivid textures.

\begin{table}[]
\centering
\setlength\extrarowheight{1pt}
\resizebox{.5\textwidth}{!}{%
\centering
\begin{tabular}{c||c|c|P{1.5cm}|P{1.5cm}}
\hline
&Sequential&Motion & \multirow{2}{*}{FID-Video}& \multirow{2}{*}{AED} \\ 
&Generation&Denoise&&  \\\hline
&&&&\\ [-8pt]\hline
Naive-Animation    & N          & N   & 4.176 & 0.0141 \\
Seq-Animation      & Y          & N   & 3.685 & 0.0128 \\
Ours-Animation     & Y          & Y   & \textbf{3.426} & \textbf{0.0126} \\ \hline
\end{tabular}%
}
\caption{The ablation study results of our image animation model.}
\label{tab:AblationStudyVideo}
\end{table}

\begin{figure*}[t]
\begin{center}
\includegraphics[width=0.96\linewidth]{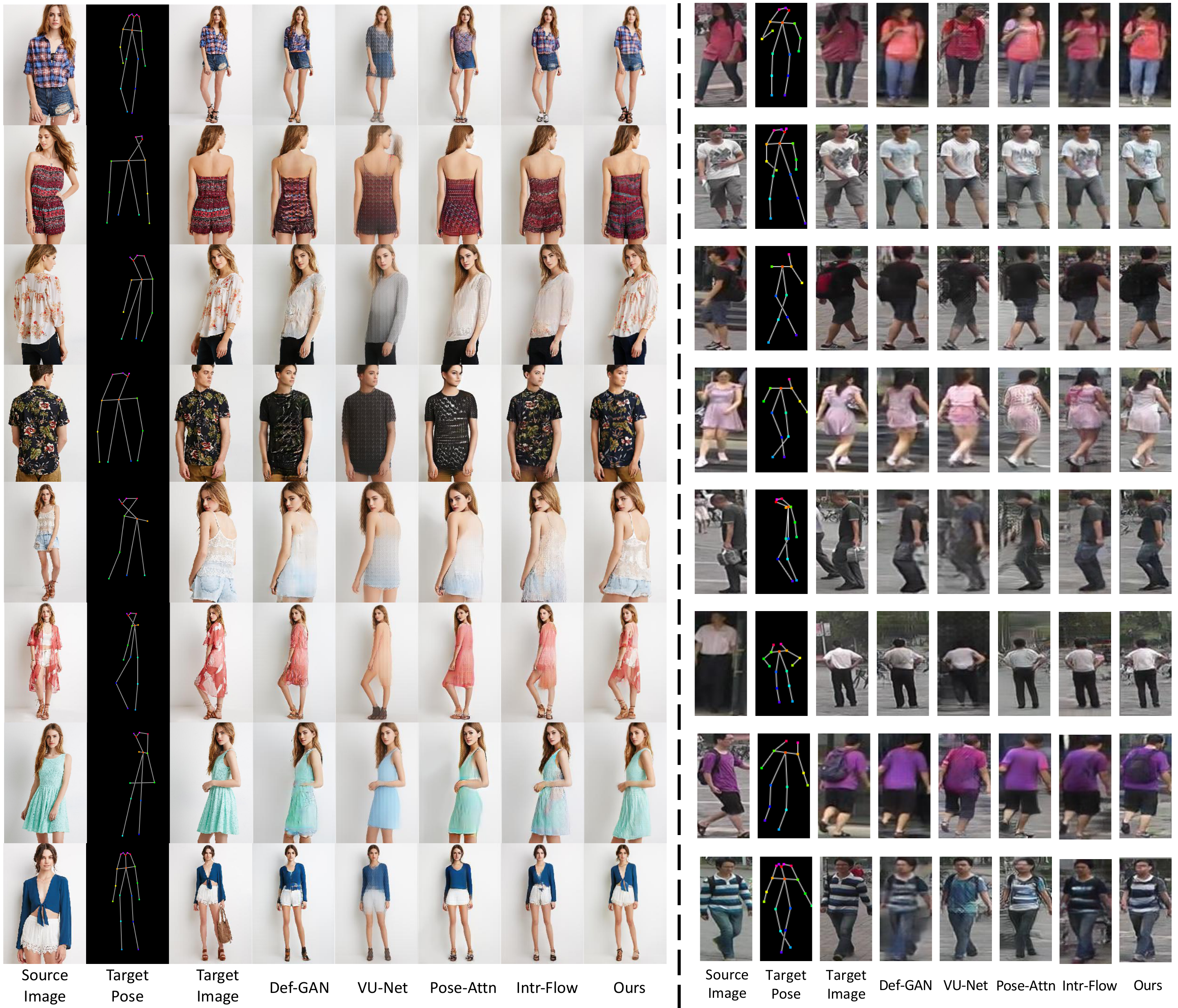}
\end{center}
\caption{Qualitative comparisons with several state-of-the-art person image generation models including Def-GAN~\cite{siarohin2018deformable}, VU-Net~\cite{esser2018variational}, Pose-Attn\cite{zhu2019progressive}, and Intr-Flow~\cite{li2019dense}. The left part shows the results of the DeepFashion dataset. The right part shows the results of the Market-1501 dataset.}
\label{fig:Compare}

\end{figure*}
\subsection{Comparisons}
\label{comparisons}

In this section, we compare our method to several state-of-the-art models on both generation and animation tasks. 
For the person image generation task, popular methods Def-GAN~\cite{siarohin2018deformable}, VU-Net~\cite{esser2018variational}, Pose-Attn\cite{zhu2019progressive} and Intr-Flow~\cite{li2019dense} are selected as the competitors. 
The quantitative evaluation results are shown in Table~\ref{tab:object}. Please note that we train the Market-1501 dataset using their original $128 \times 64$ images.
To alleviate the influence of the backgrounds on the reconstruction errors, we follow the previous work~\cite{ma2017pose} to provide the mask-LPIPS.
It can be seen that our model achieves competitive evaluation results, which means that our model can generate realistic results with fewer perceptual reconstruction errors.
Since subjective metrics have their own limitations, their results may mismatch with the actual subjective perceptions~\cite{zhang2018unreasonable}. Therefore, a human objective evaluation test is performed. A JND test is implemented on Amazon Mechanical Turk (MTurk). Volunteers are  asked to choose the more realistic image from image pairs of real and generated images. The test is performed over $800$ image pairs for each model and dataset. 
To avoid individual bias, each image pair is compared $5$ times by different volunteers. The results can be found in Table~\ref{tab:object}. It can be seen that our model achieves the best result in the challenging Fashion dataset and competitive results in the Market-1501 dataset. Besides, we provide the numbers of model parameters to evaluate the computation complexity. Thanks to our efficient spatial transformation blocks,
our model does not require a large number of convolution layers. Thus, we can achieve high performance with less than half of the parameters of the competitors.

\begin{table*}
\setlength\extrarowheight{1pt}
\centering
\resizebox{.95\textwidth}{!}{%

\begin{tabular}{p{2cm}||P{1.2cm}P{1.2cm}P{1.3cm}P{1.2cm}||P{1.2cm}P{1.2cm}P{1.3cm}P{1.2cm}||P{2cm}}\hline

& \multicolumn{4}{c||}{FashionVideo} & \multicolumn{4}{c||}{iPER}&  Number of \\ \cline{2-9} 
& FID  &LPIPS   &FID-Video &AED  &FID  &LPIPS  &FID-Video  & AED   & Parameters\\ \hline
&      &            &        &     &     &           &       &       &     \\ [-8pt]\hline
LiquidNet   &17.681 &0.0897 &5.174   &0.0184  &29.97  &0.1096        &9.212    &0.0251    & 97.45M\\ \hline
FewShot-V2V  &27.803 &0.0816 &5.096   &0.0188  &75.42  &0.2524   &8.213  &0.0232    & 97.96M \\ \hline
Ours-Animation&\textbf{14.95} &\textbf{0.0651}  &\textbf{3.426}     & \textbf{0.0126}   &\textbf{20.53}  & \textbf{0.0735} &\textbf{4.616}     & \textbf{0.0183} & \textbf{23.51M} \\ \hline

\end{tabular}%

}

\caption{Quantitative comparisons over dataset FashionVideo~\cite{zablotskaia2019dwnet} and iPER~\cite{liu2019liquid} with state-of-the-art person image animation methods including LiquidNet~\cite{liu2019liquid} and FewShot-V2V~\cite{wang2019few}. FID and LPIPS are image-based metrics. Their results indicate the quality of video frames. FID-Video and AED are calculated using video features. These metrics take the temporal distortions into consideration.}

\label{tab:object_animation}
\end{table*}

\begin{figure*}[t]
\begin{center}
\href{https://renyurui.github.io/GFLA-web/Animation_Comparison}{
\includegraphics[width=1\linewidth]{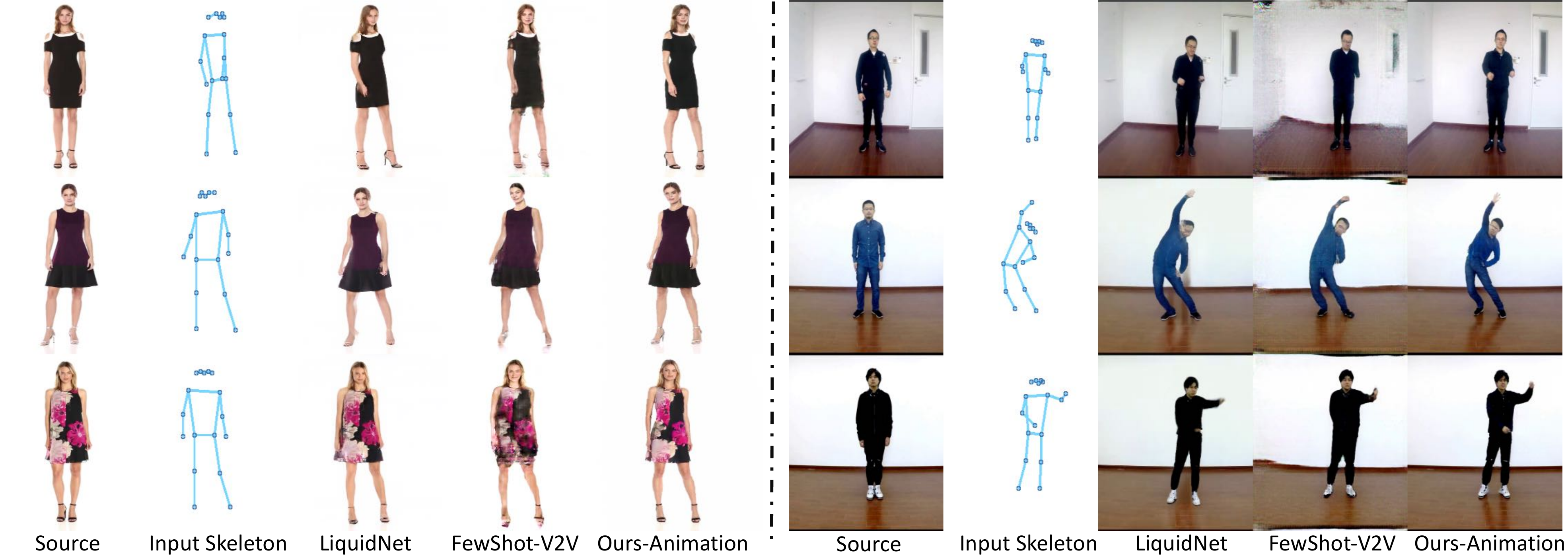}}
\end{center}
   \caption{Qualitative comparisons with person image animation models LiquidNet~\cite{liu2019liquid} and FewShot-V2V~\cite{wang2019few}.
   The left part contains the results of the FashionVideo dataset. The right part contains the results of the iPER dataset. \textit{Click on the image to play the video in a browser.}}
 
\label{fig:animation_compare}
\end{figure*}

\begin{figure}[t]
\begin{center}
\href{https://renyurui.github.io/GFLA-web/Ours_More}{
\includegraphics[width=1\linewidth]{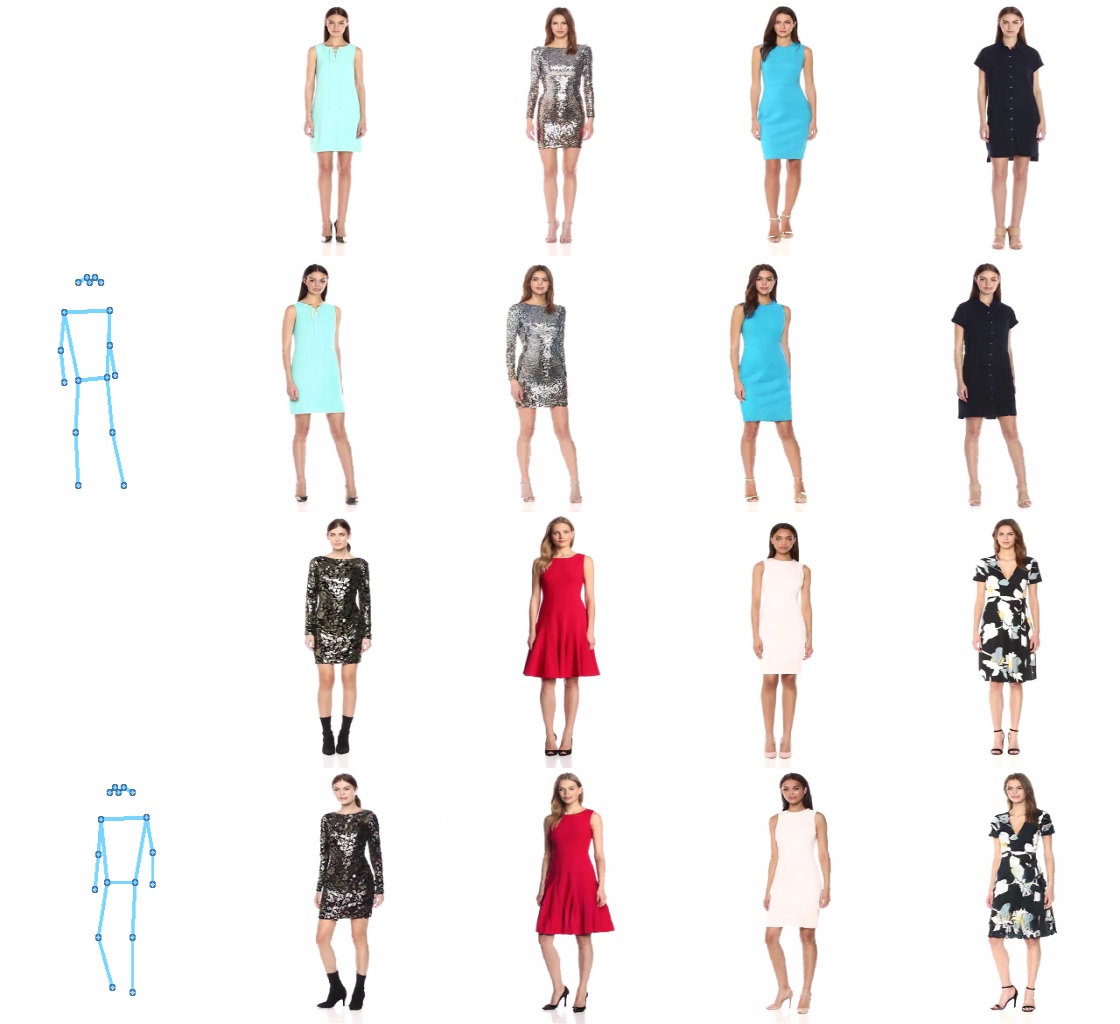}
}
\end{center}
   \caption{We show more animation results of our model. The first and third rows contain the source images. The second and fourth rows contain the driven skeleton sequences and the generated results. \textit{Click on the image to play the video in a browser.}}

\label{fig:more_results}
\end{figure}
We provide the typical results of different methods in Figure~\ref{fig:Compare}.
For the Fashion dataset, VU-Net and Pose-Attn struggle to generate complex textures since these models lack efficient spatial transformation blocks. Def-GAN generates correct appearances by transforming neural textures with pre-defined local affine transformation components (eg. arms and legs \emph{etc}.). However, the affine transformation sets are not sufficient to represent the complex spatial variance, which limits the model performance.
The flow-based model Intr-Flow is able to generate vivid textures for front pose images. 
However, it fails to generate realistic results for side pose images.
The possible explanation is that this model requires predicting 3D human meshes from 2D images to generate the training flow field labels. 
Its performance is vulnerable to 3D meshes estimation errors.
Our model does not require supplementary labels and obtains accurate flow fields in a self-supervised manner.
Thanks to our efficient deep spatial transformation module, we can well preserve the complex textures of the source images.
It can be seen that our model generates realistic results with not only correct global patterns but also the vivid details such as the lace of clothes and the shoelace.
For the Market-1501 Dataset, artifacts are observed in the results of competitors, such as the sharp edges in Pose-Attn and the halo effects in Def-GAN.
Our model is able to generate realistic images. 
However, it is worth noting that our model does not achieve significant advantages over the competitors. 
The main reason is that the low-resolution images in this dataset do not contain complex textures, which prevents our advantages from being fully utilized.





For the pose-guided animation task, we compare our model with FewShot-V2V~\cite{wang2019few} and LiquidNet~\cite{liu2019liquid}. 
The comparison results are shown in Table~\ref{tab:object_animation}. 
Different from the competitors which employ either face refine models or background inpainting models to improve their results, we do not use any post-processing methods. 
It can be seen that our model achieves the best results on both datasets.
LiquidNet achieves good FID and LPIPS scores, which means that it can generate realistic video frames. 
However, the relatively poor FID-Video and AED scores indicate that the temporal consistency is not well-maintained.
Although FewShot-V2V achieve good results on the video-based metrics, it may suffer from some image-based artifacts, which leads to poor FID scores.
Our model can generate coherent results with realistic frames. 
Meanwhile, we use significantly fewer model weights than competitors.

The subjective results are provided in Figure~\ref{fig:animation_compare}.
It can be seen that the LiquidNet model struggles to maintain temporal consistency.
This is because this model generates each frame independently.
The FewShot-V2V model solves this problem by modeling the correlations between adjacent frames.
Although this model can generate coherent results, it suffers from artifacts when generating images with complex textures or backgrounds. 
Our sequential GFLA model efficiently builds temporal dynamics. 
Meanwhile, the accurate neural texture transformation module helps with preserving the realistic details.
Therefore, our model can generate results with not only correct textures but also vivid temporal details such as the folds of clothes and the movements of hemlines. 
We provide more results of our model in Figure~\ref{fig:more_results}. It can be seen that our model is able to generate realistic videos even for source images with complex textures.







\section{Application on Other Tasks}
In this section, we show that our model is not limited to generating person images. It can be flexibly applied to other tasks requiring spatial transformation. Additional experiments are shown on two typical example tasks: novel view synthesis and face image animation.

\begin{figure}[t]
\begin{center}
\href{https://renyurui.github.io/GFLA-web/ShapeNet}{
\includegraphics[width=1\linewidth]{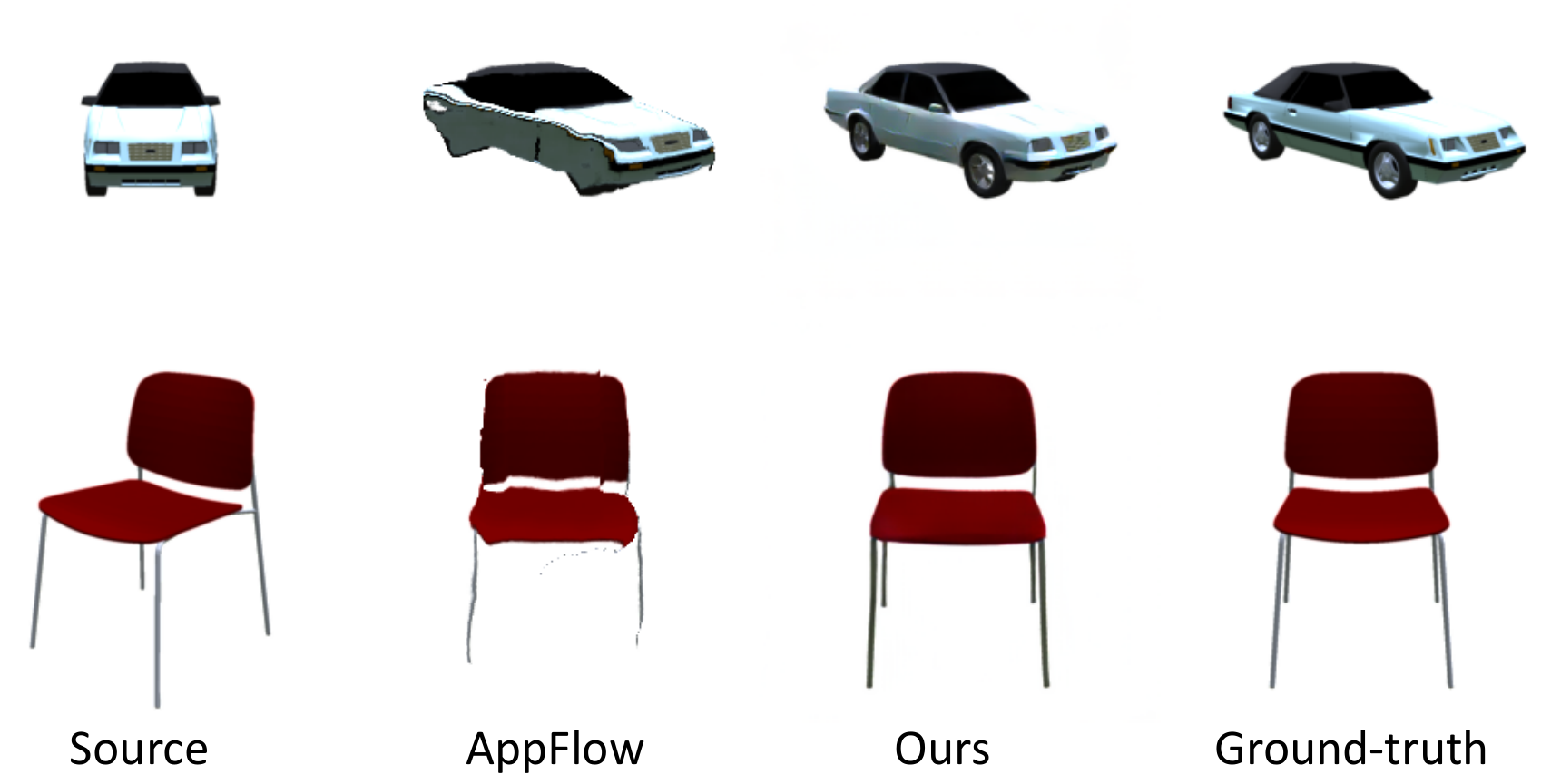}}
\end{center}
   \caption{Qualitative results of the view synthesis task. We show the results of our model and appearance flow~\cite{zhou2016view} (AppFlow) model. \textit{Click on the image to play the video in a browser.}}
 
\label{fig:shape_net}
\end{figure}

\noindent
\textbf{Novel view synthesis} requires generating new images of an object observed from arbitrary viewpoints. It can be solved by spatially transforming the source information. The car and chair categories of the ShapeNet dataset~\cite{chang2015shapenet} are used in this experiment. We train the GFLA model described in Section~\ref{image_generation}. The results can be found in Figure~\ref{fig:shape_net}. We provide the results of appearance flow~\cite{zhou2016view} which warps the source images at the pixel level as a comparison. It can be seen that appearance flow is able to transform the contents in the source images. However, it struggles to reconstruct the occluded details. Our model generates realistic images.

\begin{figure}[t]
\begin{center}
\href{https://renyurui.github.io/GFLA-web/Face_Animation}{
\includegraphics[width=1\linewidth]{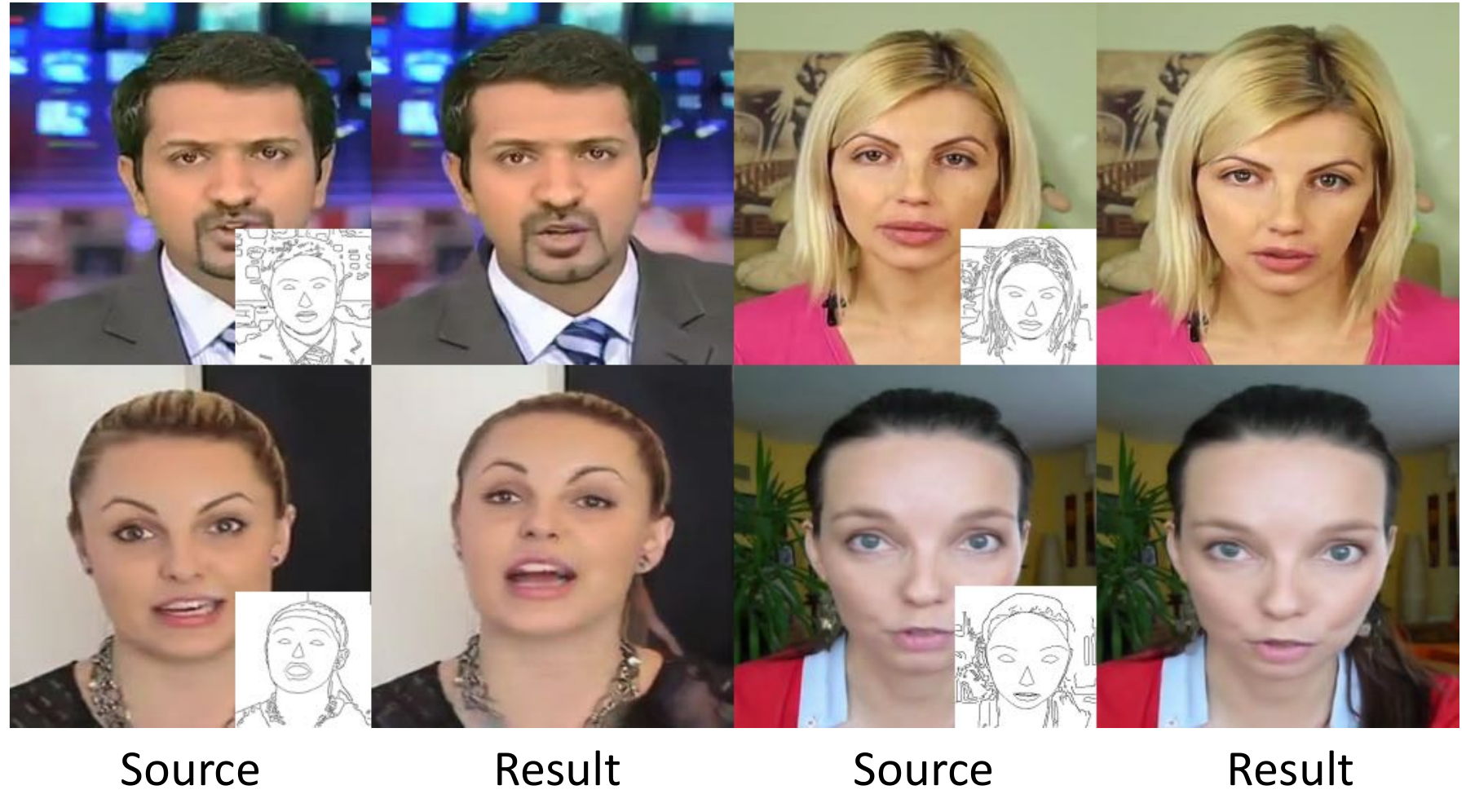}}
\end{center}
   \caption{Qualitative results of the face image animation task. \textit{Click on the image to play the video in a browser.}}
 
\label{fig:face_animation}
\end{figure}

\noindent
\textbf{Face image animation} is to generate a coherent face video clip according to a source image and a driven structure sequence. Similar to the person image animation task, this task also requires spatial manipulation of source data. We employ the real videos in the FaceForensics dataset~\cite{rossler2018faceforensics} This dataset contains $1000$ videos of news briefings from different reporters. We follow the previous papers~\cite{wang2018video,wang2019few} to use the edge maps as the structure guidance. Our sequential generator described in Section~\ref{sequential_GFLA} is employed to tackle this task. We show the qualitative results in Figure~\ref{fig:face_animation}. It can be seen that our model can generate temporally consistent results with realistic textures.







\section{Conclusion and Future Work} 
\label{sec:Conclusion}
In this paper, we tackle the person image generation and animation tasks using deep spatial transformation. 
We analyze the possible reasons causing poor gradient propagation when warping sources at the feature level. 
Targeted solution GFLA framework is proposed to first estimate flow fields between sources and targets and then sample the source features in a content-aware manner.
We have demonstrated empirically that the GFLA model can provide improved gradients, leading to accurate spatial transformations. 
Meanwhile, we further propose a sequential GFLA model to extract the correlations between adjacent frames for the animation task. Experiments show that our model can efficiently build temporal dynamics and generate coherent videos. Finally, we demonstrate that our model is versatile on other tasks requiring spatial transformation such as face image animation and novel view synthesis. 

\begin{figure}[h!]
\begin{center}
{\includegraphics[width=1\linewidth]{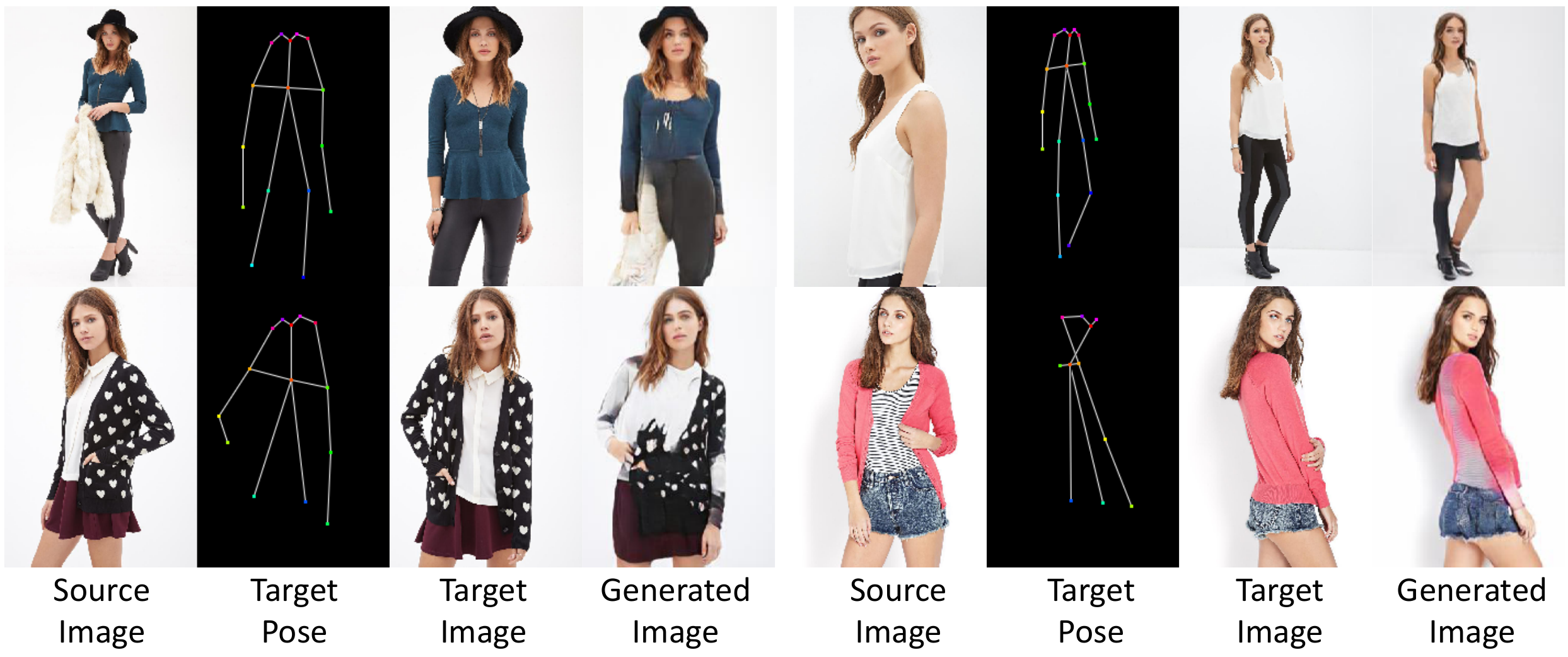}}
\end{center}
   \caption{Some failure cases of our GFLA model.}
 
\label{fig:failure_cases}
\end{figure}

Although our model generates impressive results, we also observe some failure cases as shown in Figure~\ref{fig:failure_cases}. 
These typical failure cases are due to the severe occlusions of source images, which misleads the model to sample incorrect neural textures.
We provide possible solutions for this open issue to inspire future works in this problem.
One way is to add additional constraints to flow fields. For example, loss functions can be designed to penalize sampling between different semantic regions. 
Another solution is to perform multi-step warping operations to gradually warp source images to targets by using additional video datasets.

\ifCLASSOPTIONcaptionsoff
  \newpage
\fi



\bibliographystyle{IEEEtran}
\bibliography{egbib}
%


%

\begin{IEEEbiography}[{\includegraphics[width=1in,height=1.25in,clip,keepaspectratio]{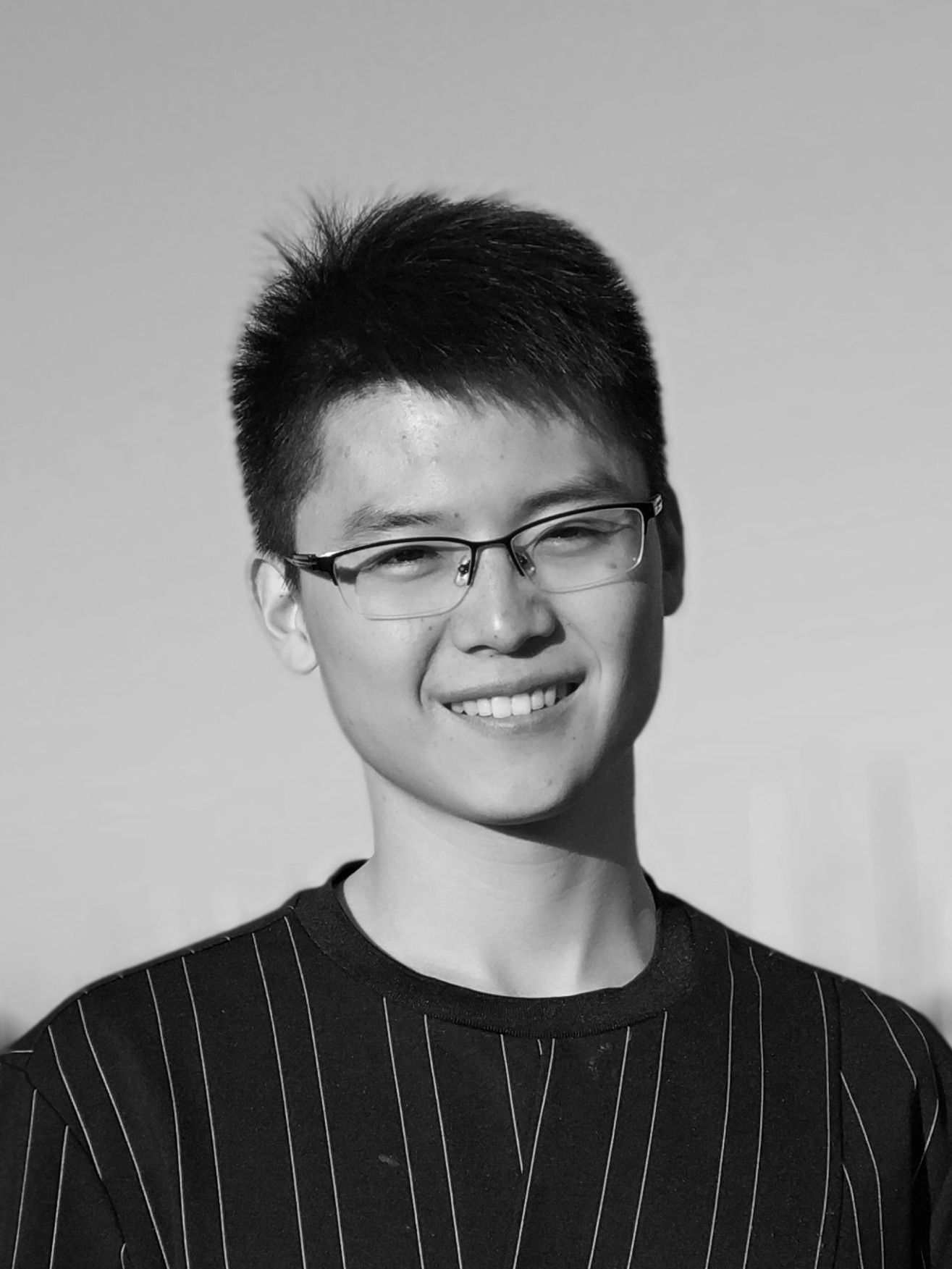}}]{Yurui Ren}
received the B.S. degree form School of Electronic Information and Electrical Engineering, Dalian University of Technology, Dalian, China, in 2017. He is currently pursuing the Ph.D. degree in School of Electronic and Computer Engineering, Peking University, Shenzhen, China. His research interests include image generation and image enhancement.
\end{IEEEbiography}

\begin{IEEEbiography}[{\includegraphics[width=1in,height=1.25in,clip,keepaspectratio]{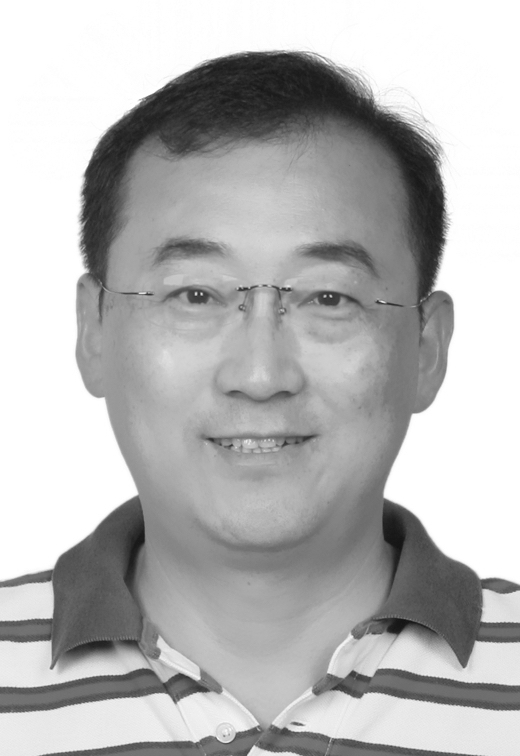}}]{Ge Li}
is a professor at the School of Electronic and Computer Engineering in Peking University Shenzhen Graduate School, China. His research interests include image/video process and analysis, machine learning, digital communications and signal processing.
\end{IEEEbiography}

\begin{IEEEbiography}[{\includegraphics[width=1in,height=1.25in,clip,keepaspectratio]{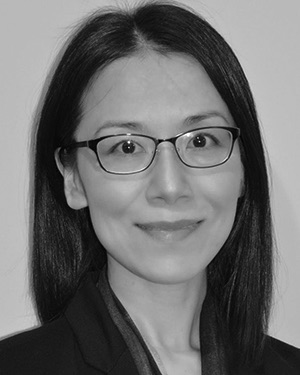}}]{Shan Liu}
received the B.Eng. degree in electronic engineering from Tsinghua University, the M.S. and Ph.D. degrees in electrical engineering from the University of Southern California, respectively.

She is a Tencent Distinguished Scientist and General Manager of Tencent Media Lab. She was formerly Director of Media Technology Division at MediaTek USA. She was also formerly with MERL, Sony and IBM. She has been actively contributing to international standards since the last decade and served co-Editor of HEVC SCC and the emerging VVC. She has numerous technical contributions adopted into various standards, such as HEVC, VVC, OMAF, DASH and PCC, etc. At the same time, technologies and products developed by her and under her leadership have served several hundred million users. Dr. Liu holds more than 150 granted US and global patents and has published more than 80 journal and conference papers. She was in the committee of Industrial Relationship of IEEE Signal Processing Society (2014-2015) and is on the Editorial Board of IEEE Transactions on Circuits and Systems for Video Technology (2018-2021). She was the VP of Industrial Relations and Development of Asia-Pacific Signal and Information Processing Association (2016-2017) and was named APSIPA Industrial Distinguished Leader in 2018. She was appointed Vice Chair of IEEE Data Compression Standards Committee in 2019. Her research interests include audio-visual, high volume, immersive and emerging media compression, intelligence, transport, and systems.
\end{IEEEbiography}

\begin{IEEEbiography}[{\includegraphics[width=1in,height=1.25in,clip,keepaspectratio]{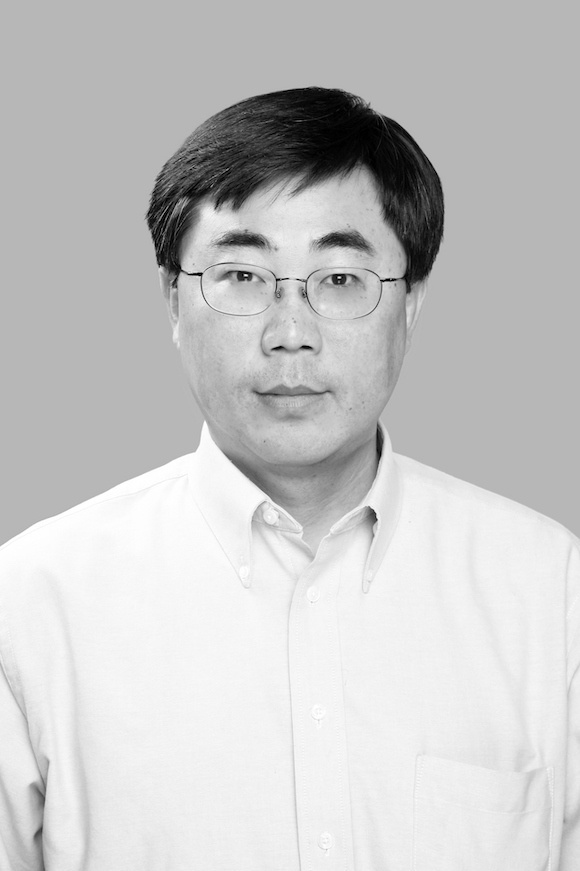}}]{Thomas H. Li}

received the BE in electronics engineering from Beijing Institute of Technology, Beijing, China, in 1982, and the MSEE and Ph.D. in electrical engineering from Purdue University, West Lafayette, Indiana, USA in 1991 and 1999, respectively.  
After working in industry for over 20 years, he joined Gpower Semiconductor, Inc., in Suzhou, China, as the Chief Strategist. His research interests include communication, signal processing and machine learning. 
\end{IEEEbiography} 
 






\clearpage
\onecolumn
\newcounter{alphasect}
\def\alphainsection{0}

\let\oldsection=\section
\def\section{%
  \ifnum\alphainsection=1%
    \addtocounter{alphasect}{1}
  \fi%
\oldsection}%

\renewcommand\thesection{%
  \ifnum\alphainsection=1%
    \Alph{alphasect}
  \else%
    \arabic{section}
  \fi%
}%

\newenvironment{alphasection}{%
  \ifnum\alphainsection=1%
    \errhelp={Let other blocks end at the beginning of the next block.}
    \errmessage{Nested Alpha section not allowed}
  \fi%
  \setcounter{alphasect}{0}
  \def\alphainsection{1}
}{%
  \setcounter{alphasect}{0}
  \def\alphainsection{0}
}%
\begin{alphasection}

\section{\textbf{Additional Results of Person Image Generation}}
We provide additional comparisons with state-of-the-art person image generation models in this section. The qualitative results is shown in Figure~\ref{fig:compare}.

\begin{figure}[H]
\renewcommand\thefigure{\Alph{section}.\arabic{figure}}
\begin{center}
\includegraphics[width=0.97\linewidth]{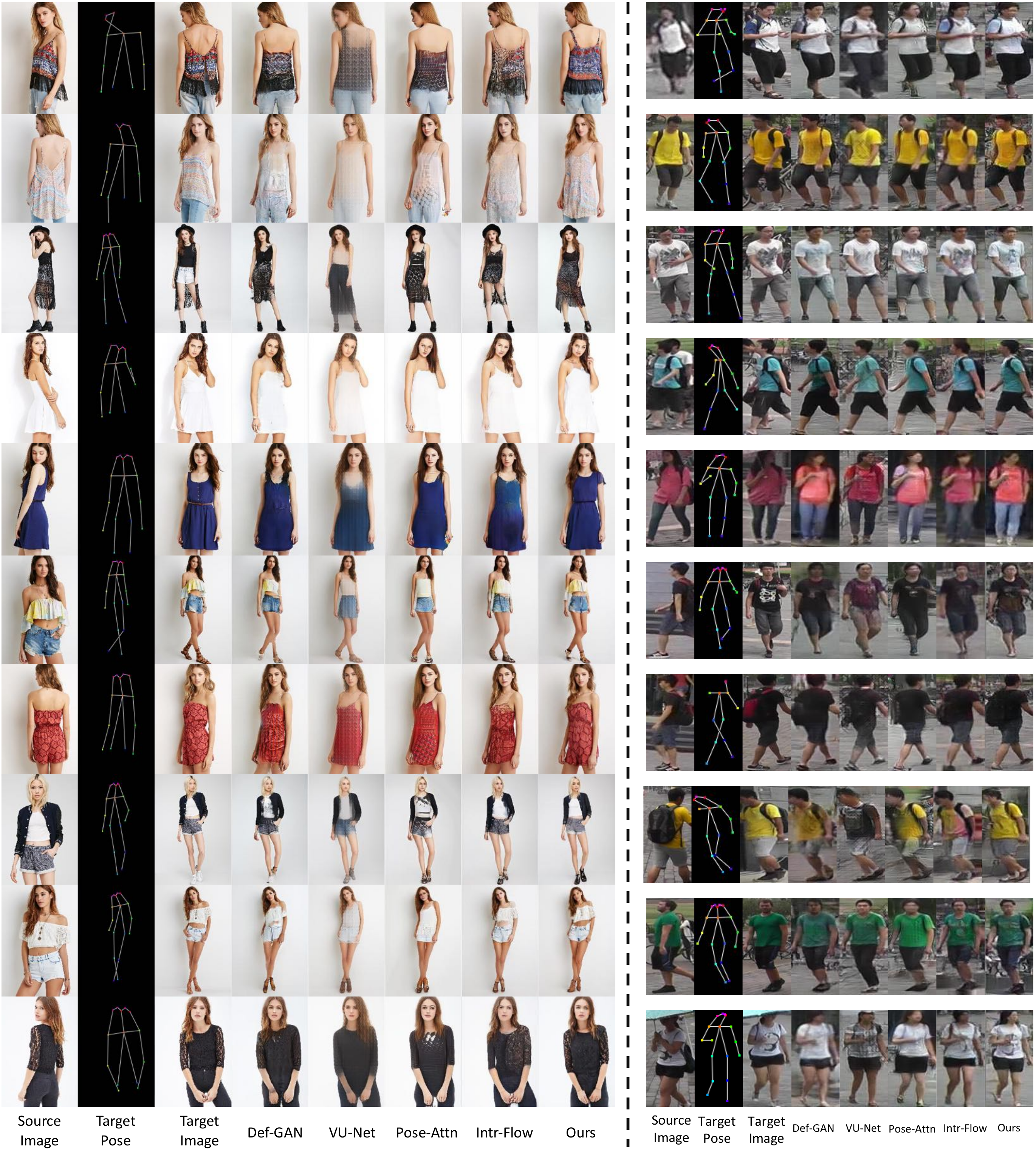}
\end{center}
\caption{The qualitative comparisons with several state-of-the-art models including Def-GAN~\cite{siarohin2018deformable}, VU-Net~\cite{esser2018variational}, Pose-Attn\cite{zhu2019progressive}, and Intr-Flow~\cite{li2019dense} over dataset DeepFashion~\cite{liu2016deepfashion} and Market-1501~\cite{zheng2015scalable}.}
\label{fig:compare}
\end{figure}

\clearpage
\section{\textbf{Additional Results of Person Image Animation}}
\label{sec:animation}
We provide additional results of the person image animation task in Figure~\ref{fig:person_animation}.
\begin{figure}[ht]
    \renewcommand\thefigure{\Alph{alphasect}.\arabic{figure}}

    \offinterlineskip

    \centering
    \href{https://renyurui.github.io/GFLA-web/Additional_Animation_Comparison}{
    \includegraphics[width=0.95\linewidth]{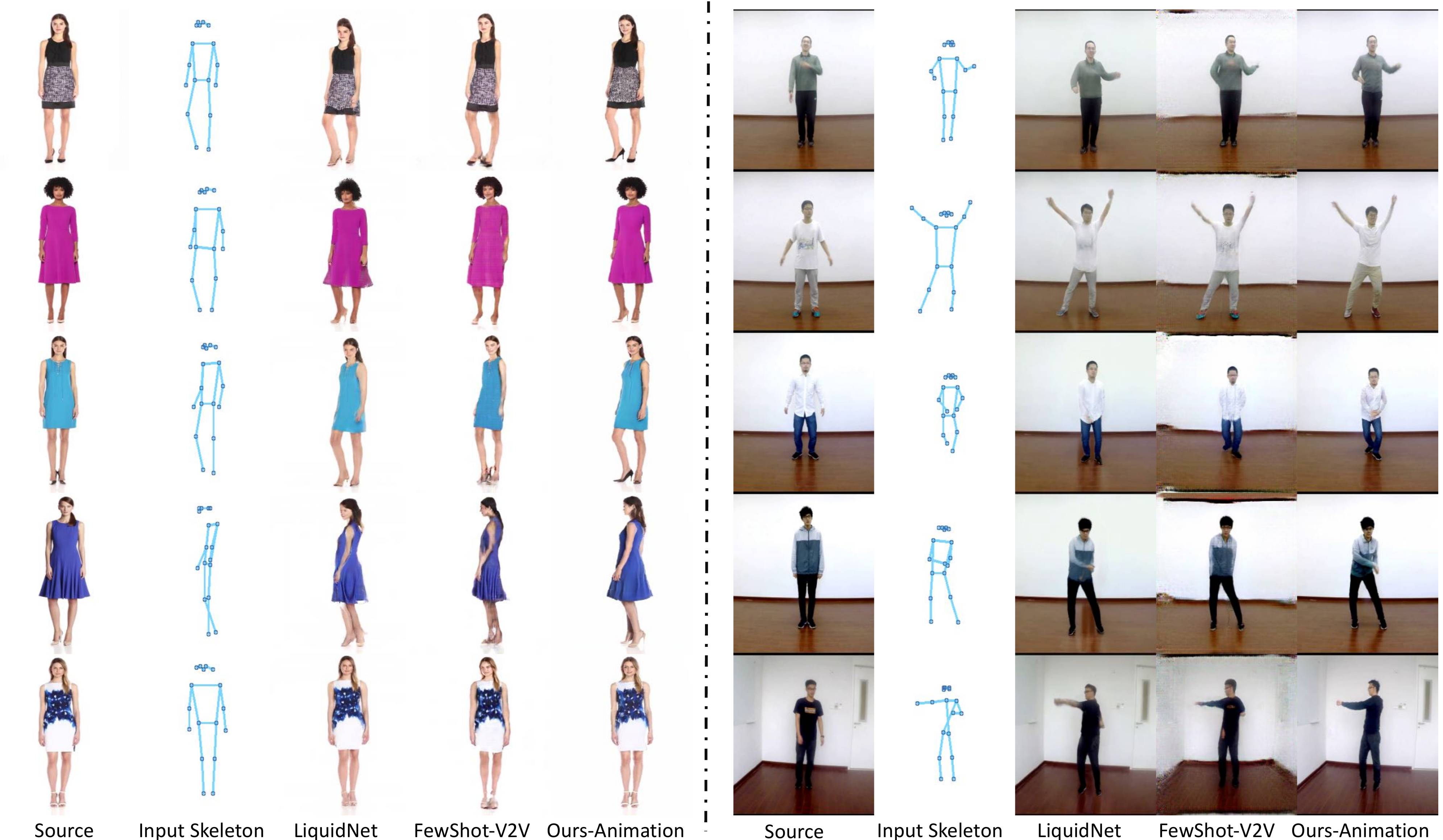}}
 
    \caption{Qualitative results of the person image animation task. We compare our model with stare-of-the-art person image animation models including LiquidNet~\cite{liu2019liquid} and FewShot-V2V~\cite{wang2019few}. \textit{Click on the image to start the animation in a browser}.}
    \label{fig:person_animation}
\end{figure}

\section{\textbf{Additional Results of Face Image Animation}}
We provide additional results of the face image animation task in Figure~\ref{fig:face}.
\begin{figure}[ht]
    \renewcommand\thefigure{\Alph{alphasect}.\arabic{figure}}

    \offinterlineskip

    \centering
    \href{https://user-images.githubusercontent.com/30292465/75650849-1d03f280-5c92-11ea-9f3d-ae4c85524787.gif}{
    \includegraphics[width=0.95\linewidth]{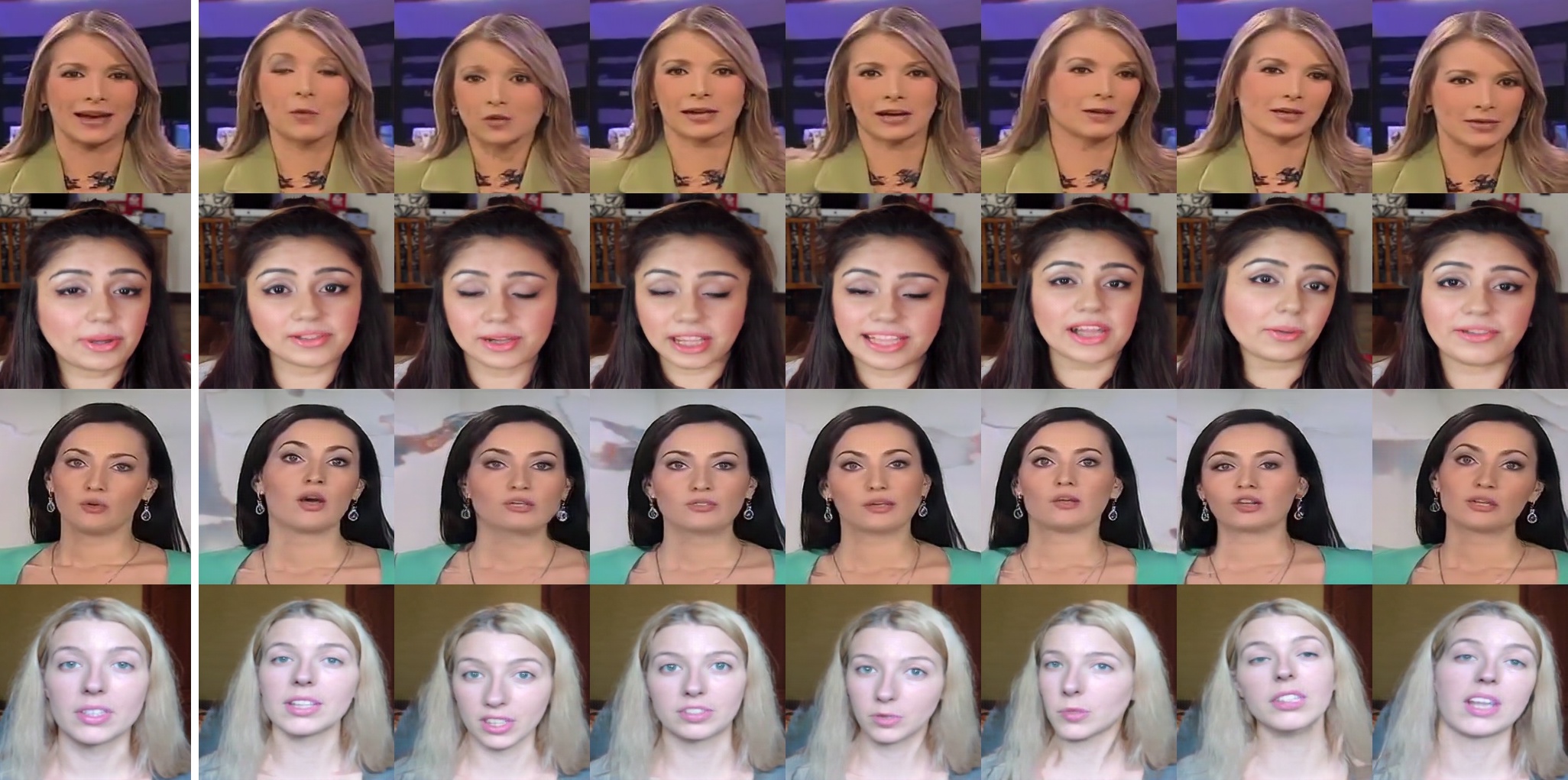}}
 
    \caption{Qualitative results of the image animation task. For each row, the leftmost image is the source image. The others are generated images. \textit{Click on the image to start the animation in a browser}.}
    \label{fig:face}
\end{figure}

\clearpage
\section{Additional Results of View Synthesis}
\label{sec:view}
We provide additional results of the view synthesis task in Figure~\ref{fig:car} and Figure~\ref{fig:chair}.
\begin{figure}[H]
    \renewcommand\thefigure{\Alph{alphasect}.\arabic{figure}}
    \offinterlineskip
 
    \centering
    \href{https://user-images.githubusercontent.com/30292465/75650787-f8a81600-5c91-11ea-998e-94685f956302.gif}{
    \includegraphics[width=1\linewidth]{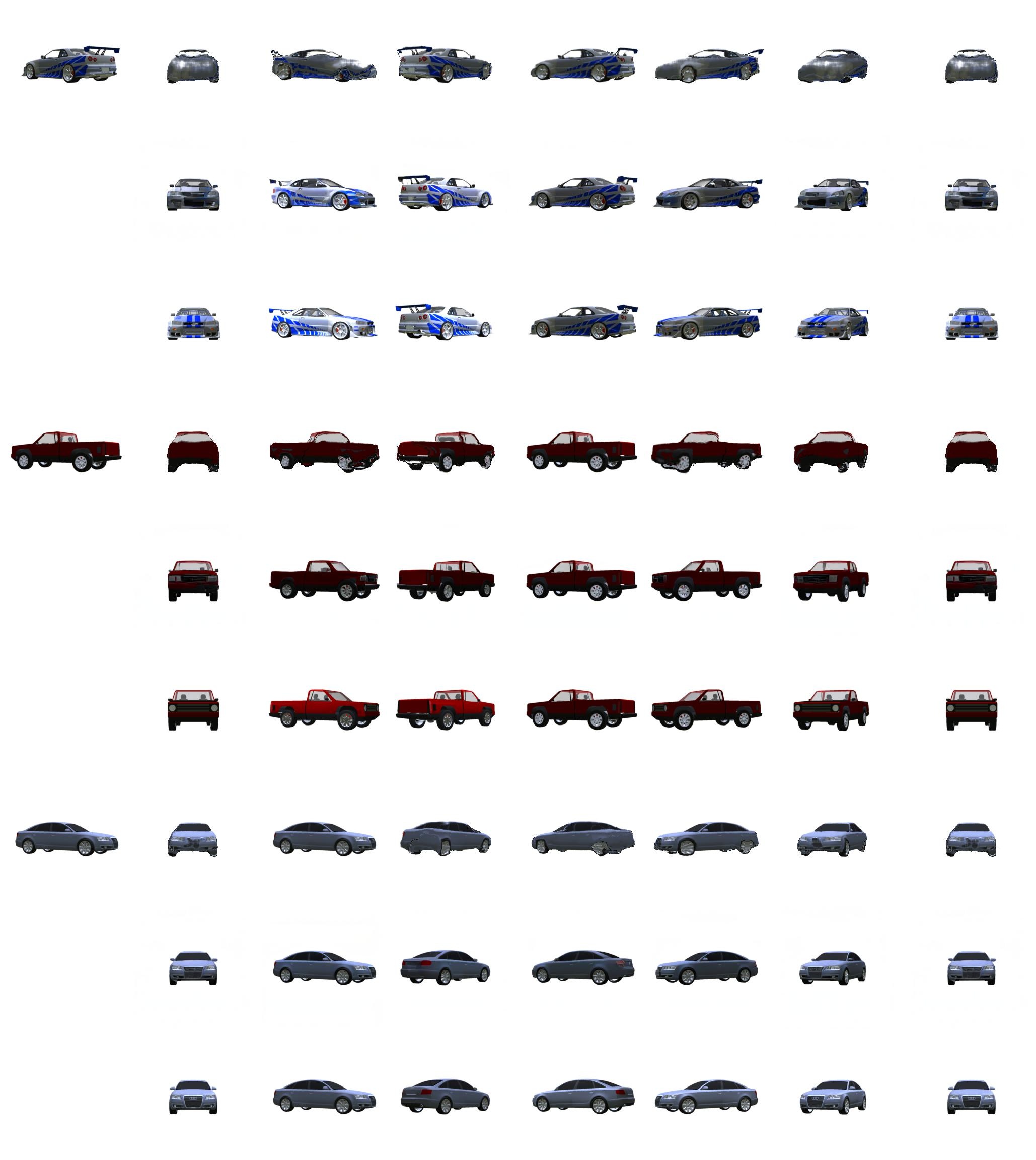}}

    \caption{Qualitative results of the view synthesis task. For each group, we show the results of Appearance Flow~\cite{zhou2016view}, the results of our model, and ground-truth images, respectively. The top left image is the input source image. The other images are the generated results and ground-truth images. \textit{Click on the  image to start the animation in a browser}.}
    \label{fig:car}
\end{figure}

\begin{figure}[H]
    \renewcommand\thefigure{\Alph{alphasect}.\arabic{figure}}

    \offinterlineskip

    \centering
    \href{https://user-images.githubusercontent.com/30292465/75650814-09588c00-5c92-11ea-8c08-52662312c81d.gif}{
    \includegraphics[width=1\linewidth]{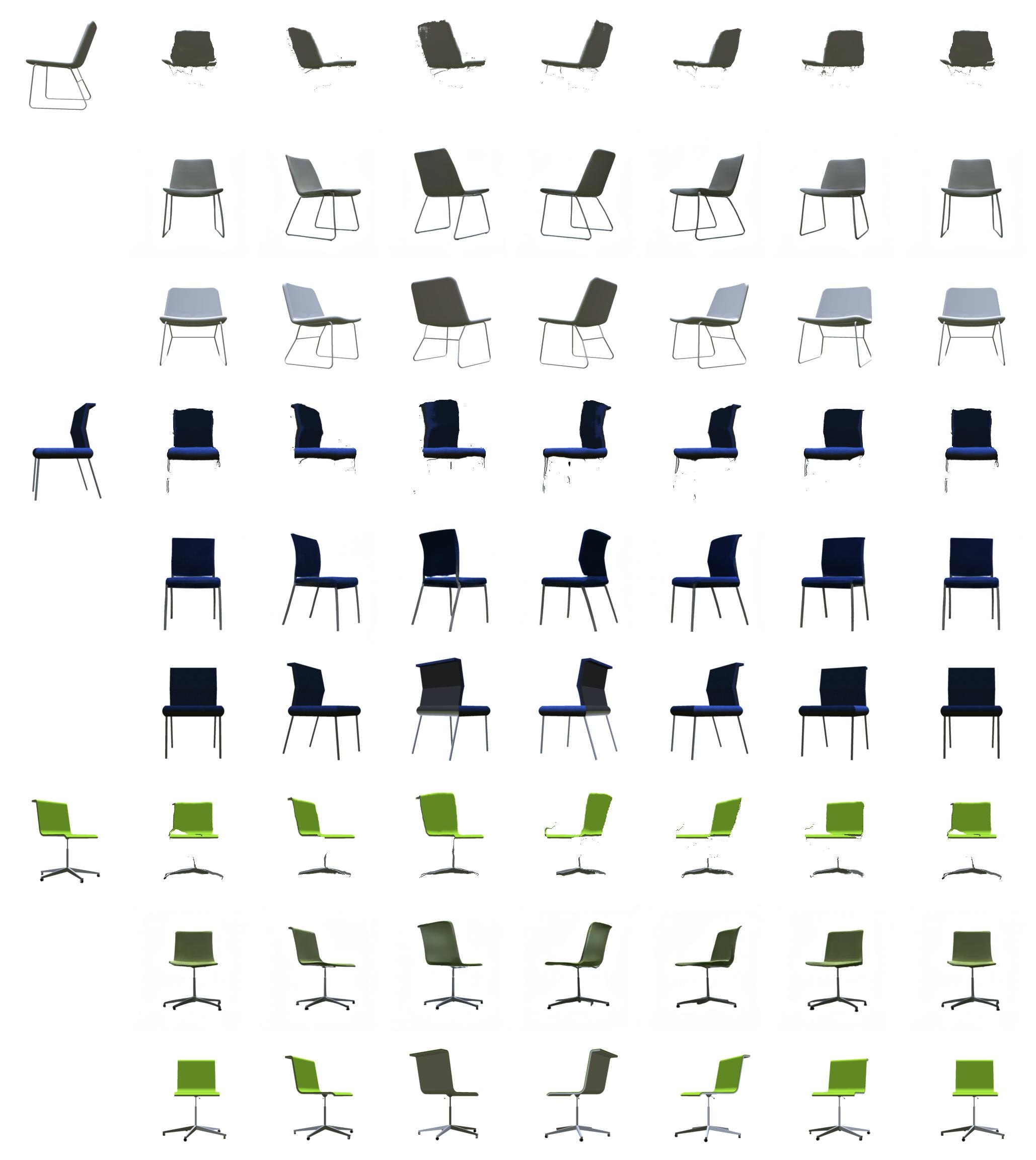}}

    \caption{Qualitative results of the view synthesis task. For each group, we show the results of Appearance Flow~\cite{zhou2016view}, the results of our model, and ground-truth images, respectively. The top left image is the input source image. The other images are the generated results and ground-truth images. \textit{Click on the image to start the animation in a browser}.}
    \label{fig:chair}
\end{figure}

\clearpage

\section{Implementation Details}
\label{sec:implementation_detail}
Basically, the auto-encoder structure is employed to design our networks. We use the residual blocks as shown in Figure~\ref{fig:network_component} to build our model. Each convolutional layer is followed by instance normalization~\cite{ulyanov2016instance}. We use Leaky-ReLU as the activation function in our model. Spectral normalization~\cite{miyato2018spectral} is employed in the discriminator to solve the notorious problem of instability training of generative adversarial networks. 

For our GFLA model, we show the architecture in Figure~\ref{fig:network}. We note that since the images of the Market-1501 dataset are low-resolution images ($128 \times 64$), we only use one local attention block at the feature maps with resolution as $32 \times 16$. 
We design the kernel prediction net $M$ in the local attention block as a fully connected network. The extracted local patch $\mathcal{N}_n(\mathbf{f}_s,l+\mathbf{w}^l)$ and $\mathcal{N}_n(\mathbf{f}_t,l)$ are concatenated as the input. The output of the network is $\mathbf{k}_l$.
Since it needs to predict attention kernels $\mathbf{k}_l$ for all location $l$ in the feature maps, we use a convolutional layer to implement this network, which can take advantage of the parallel computing power of GPUs.  
We train this model in stages. The Flow Field Estimator is first trained to generate flow fields. Then we train the whole model in an end-to-end manner. We adopt the ADAM optimizer. The learning rate of the generator is set to $10^{-4}$. The discriminator is trained with a learning rate of one-tenth of that of the generator. The batch size is set to 8 for all experiments. The loss weights are set to $\lambda_c=5$, $\lambda_r=0.0025$, $\lambda_{\ell_1}=5$, $\lambda_a=2$, $\lambda_p=0.5$, and $\lambda_s=500$.

For our person image animation model, we show the architecture of the Motion Extraction Network~\ref{fig:motion_extraction} and sequential GFLA network~\ref{fig:animation_network}. The Motion Extraction Network is designed using a similar structure as that of the paper~\cite{pavllo20193d}. We use the 1D convolutional layers as the basic component. The ADALN is used as the normalization layer. Let $\mathbf{f} \in \mathbb{R}^{N\times C \times L}$ denotes the activations of a 1D convolution layer. The ADALN normalize the inputs as

\begin{equation}
  ADALN(\mathbf{f})=\gamma(\frac{\mathbf{f}-\mu(\mathbf{f})}{\sigma(\mathbf{f})})+\beta
\end{equation}
where $\mu(\mathbf{f})$ and $\sigma(\mathbf{f})$ are computed across spatial and channel dimensions for each training case
\begin{equation}
 \mu_b(\mathbf{f}) = \frac{1}{CL}\sum_{c=1}^{C}\sum_{l=1}^{L}\mathbf{f}_{bcl} 
\end{equation}
\begin{equation}
  \sigma_b({\mathbf{f}}) = \sqrt[\leftroot{-2}\uproot{2}]{\frac{1}{CL}\sum_{c=1}^{C}\sum_{l=1}^{L}(\mathbf{f}_{bcl}-\mu_b(\mathbf{f}))^2}
\end{equation}

Instead of learning a single set of affine parameters $\gamma$ and $\beta$, we follow previous methods~\cite{huang2017arbitrary} to calculate them for each training case using the input joints. This operation allows the network to recover the input statistics (\emph{i.e.} locations, scales).
\begin{equation}
  \beta, \gamma = E(\mathbf{J}_t^{[1,K]})
\end{equation}
where $E$ is the statistic extraction module. As shown in Figure~\ref{fig:animation_network}, the sequential GFLA model has a similar architecture with that of the GFLA model. We add another path to transform the information of the previously generated images. We first train the Motion Extraction Network using the Human3.6M dataset. We use the Alphapose model~\cite{alp2018densepose} extract the noisy input skeletons. The corresponding ground-truth skeletons are provided by the dataset. After training the Motion Extraction Network, we can perprocess the skeletons of the person image animation datasets FashionVideo~\cite{li2018crowdpose} and iPER~\cite{liu2019liquid}. Finally, we train the sequential GFLA model in an end-to-end manner. For the first frame, we use the source image as the previously generated image. We adopt the ADAM optimizer. The learning rate of the generator is set to $10^{-4}$. The discriminator is trained with a learning rate of one-tenth of that of the generator. The batch size is set to 2 for all experiments.

\clearpage
\begin{figure}[t]
\renewcommand\thefigure{\Alph{alphasect}.\arabic{figure}}
\begin{center}
\includegraphics[width=0.9\linewidth]{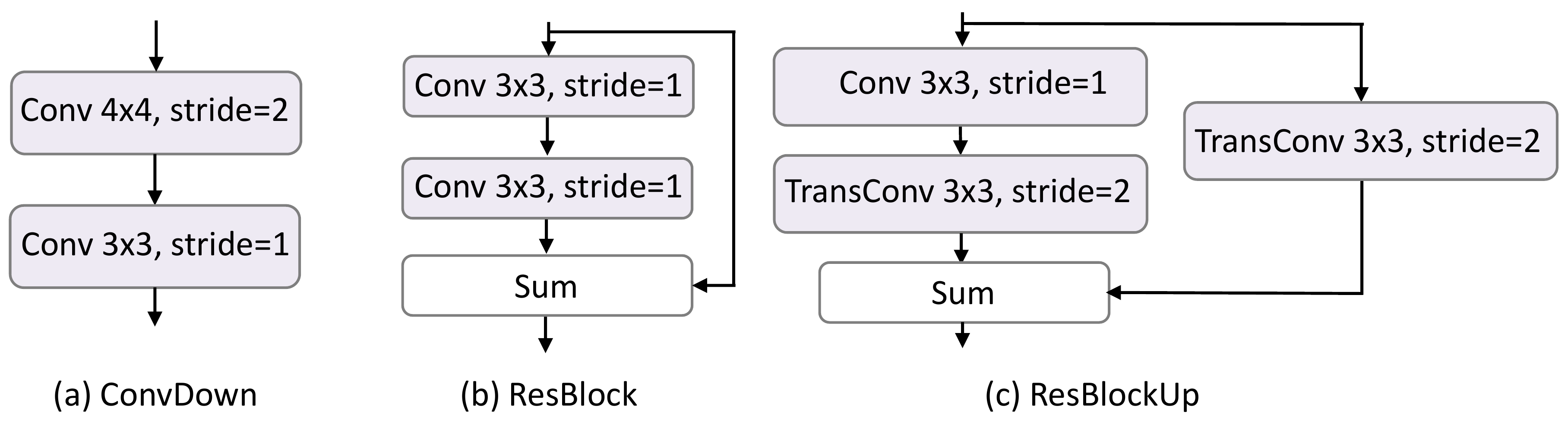}
\end{center}
\caption{The components used in our networks.}
\label{fig:network_component}
\end{figure}

\begin{figure}[b]
\renewcommand\thefigure{\Alph{alphasect}.\arabic{figure}}
\begin{center}
\includegraphics[width=0.9\linewidth]{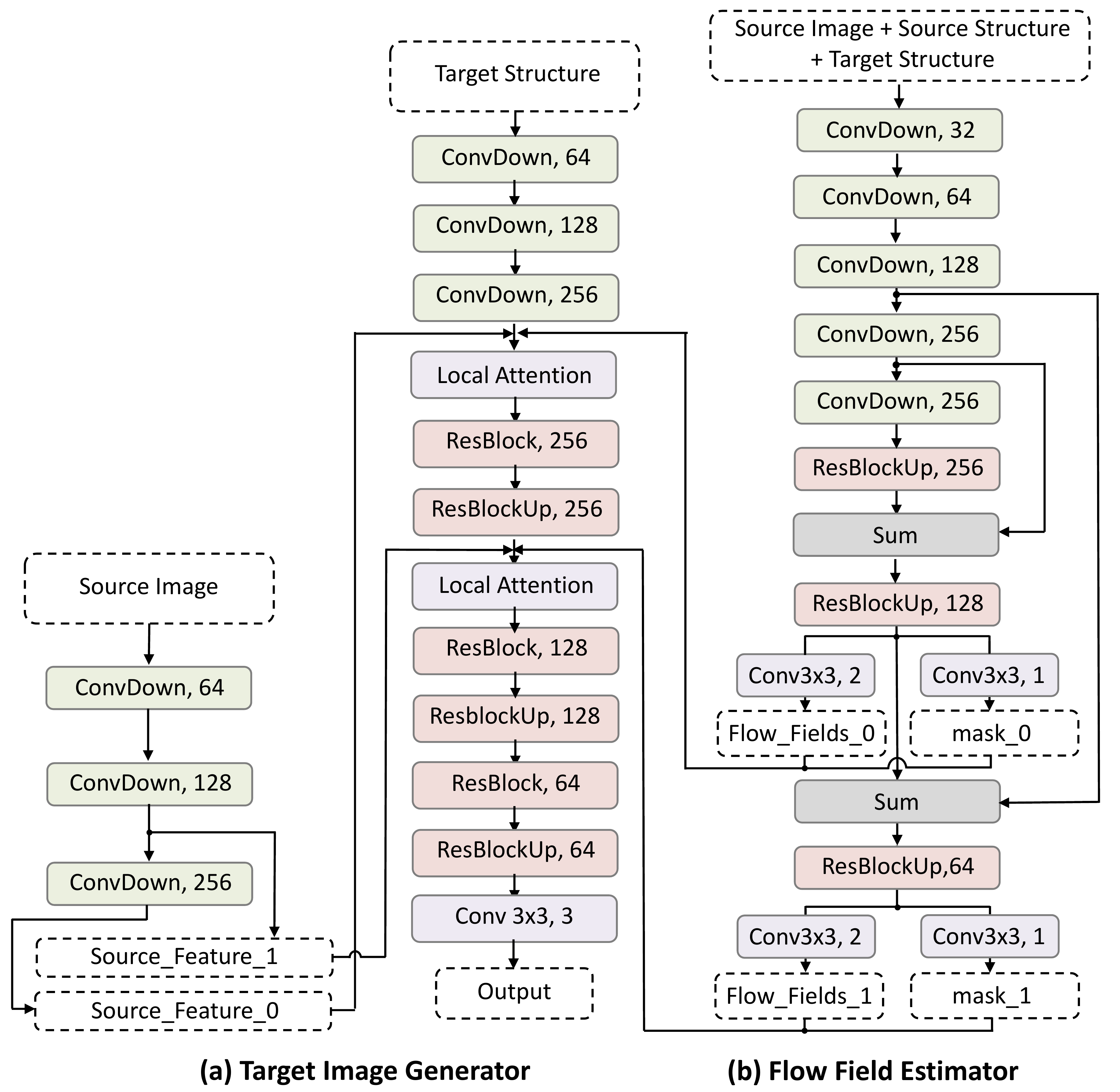}
\end{center}
\caption{The network architecture of our GFLA model.}
\label{fig:network}

\end{figure}

\clearpage

\begin{figure}[t]
\renewcommand\thefigure{\Alph{alphasect}.\arabic{figure}}
\begin{center}
\includegraphics[width=1\linewidth]{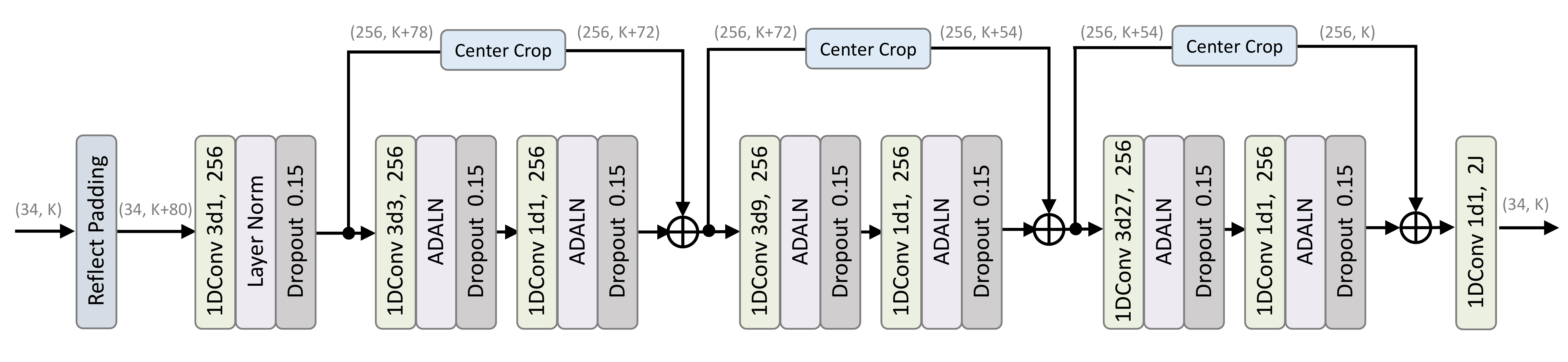}
\end{center}
\caption{The architecture of our Motion Extraction Network.}
\label{fig:motion_extraction}
\end{figure}

\begin{figure}[H]
\renewcommand\thefigure{\Alph{alphasect}.\arabic{figure}}
\begin{center}
\includegraphics[width=1\linewidth]{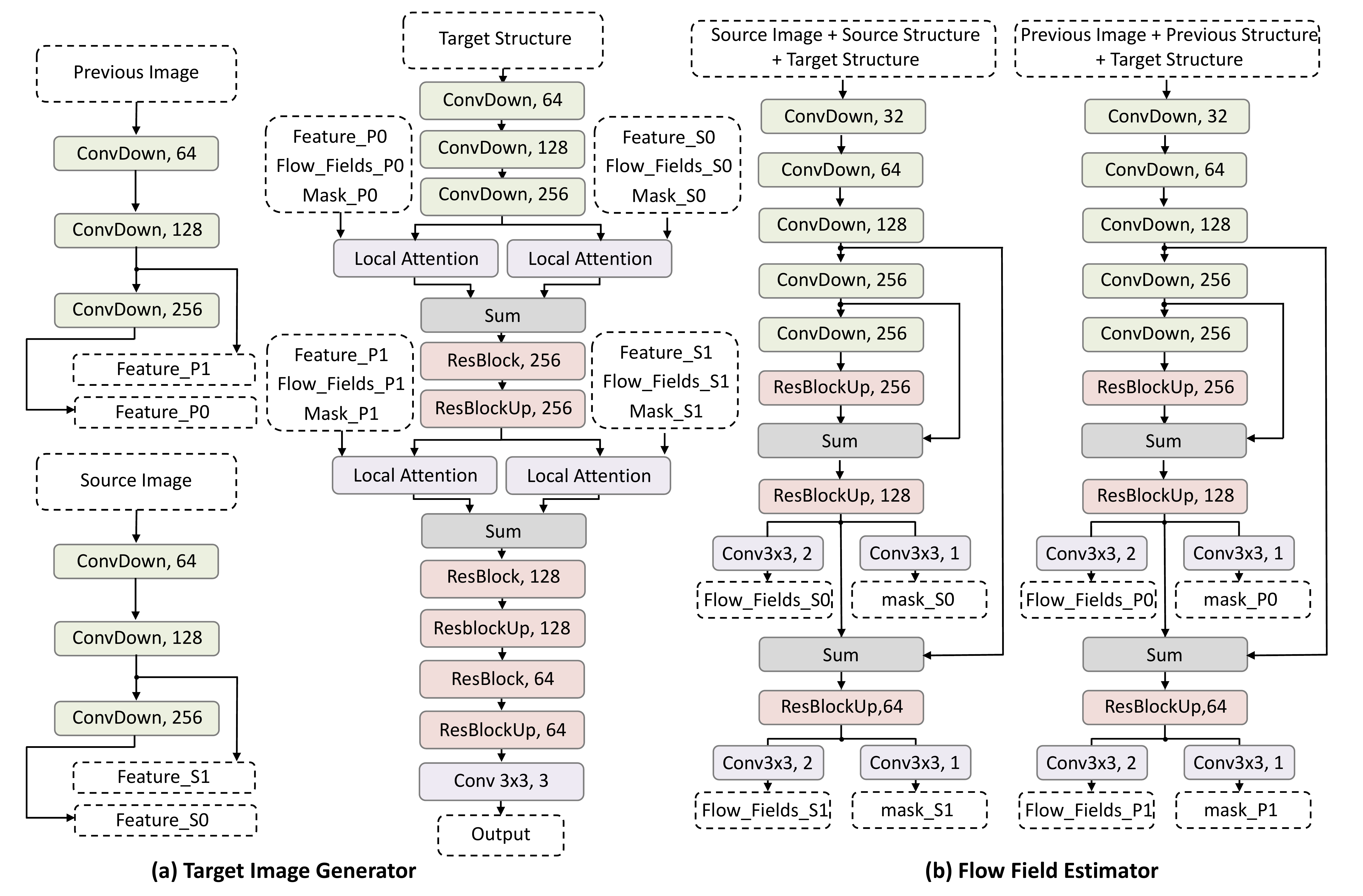}
\end{center}
\caption{The network architecture of our sequential GFLA model.}
\label{fig:animation_network}

\end{figure}
\clearpage

\end{alphasection}

\end{document}